\pdfoutput=1
\documentclass[11pt]{article}
\usepackage[final]{acl}

\usepackage{times}
\usepackage{latexsym}

\usepackage[T1]{fontenc}

\usepackage[utf8]{inputenc}

\usepackage{microtype}

\usepackage{inconsolata}

\usepackage{graphicx}

\usepackage{kotex}
\usepackage{amsmath}
\usepackage{amssymb}
\usepackage{multirow}
\usepackage{adjustbox}
\usepackage{makecell}
\usepackage{booktabs}
\usepackage[table,xcdraw]{xcolor}  % table_02
\usepackage[normalem]{ulem}        % table_02'
\useunder{\uline}{\ul}{}           % table_02'
\usepackage{twemojis}
\usepackage{algorithm} %% pseudo code
\usepackage{algpseudocode}
\usepackage{xcolor} %% color caption
\usepackage{caption}
\usepackage[normalem]{ulem}  %% table_a01
\useunder{\uline}{\ul}{}
\usepackage[table,xcdraw]{xcolor}
\usepackage{tablefootnote}
\usepackage{array}  %% table_b04
\usepackage{float}  %% figure_03,04
\usepackage{subcaption}
\usepackage{tabularx}  %% table_case_study
\usepackage{enumitem}  %% prompt
\usepackage{ragged2e}  %% prompt

\title{From Ground Trust\textsuperscript{\dag} to Truth: Disparities in Offensive Language Judgments on Contemporary Korean Political Discourse}

\author{Seunguk Yu, Jungmin Yun, Jinhee Jang \and Youngbin Kim \\
  Chung-Ang University, Seoul, Republic of Korea \\
  \texttt{seungukyu@gmail.com, \{cocoro357, jinheejang, ybkim85\}@cau.ac.kr} \\
}

\begin{document}
\maketitle
\begin{abstract}
\textit{Warning: this paper contains expressions that may offend the readers.} \\
\\
Although offensive language continually evolves over time, even recent studies using LLMs have predominantly relied on outdated datasets and rarely evaluated the generalization ability on unseen texts. In this study, we constructed a large-scale dataset of contemporary political discourse and employed three refined judgments in the absence of ground truth. Each judgment reflects a representative offensive language detection method and is carefully designed for optimal conditions. We identified distinct patterns for each judgment and demonstrated tendencies of label agreement using a leave-one-out strategy. By establishing pseudo-labels as ground trust\textsuperscript{\dag} for quantitative performance assessment, we observed that a strategically designed single prompting achieves comparable performance to more resource-intensive methods. This suggests a feasible approach applicable in real-world settings with inherent constraints.
\end{abstract}

\section{Introduction}
Offensive language has emerged as a persistent linguistic issue in online discourse, broadly encompassing expressions intended to insult, demean, or ridicule others\footnote{This study examines a broader spectrum of \textit{offensive language}, including \textit{hate speech}—expressions that promote hatred toward specific individuals or groups.}~\cite{mnassri2024survey, pradhan2020review}. This concern is particularly evident in social media comments, where users articulate and exchange diverse opinions~\cite{abdelsamie2024comprehensive, aklouche2024offensive}. The form and content of such language frequently depend on the underlying factual context~\cite{GHENAI2025104079}, and they are recognized as key factors in intensifying social tensions and driving the polarization of public opinions~\cite{vasist2024polarizing, kaur2024deep}.

In this study, we focus on Korean online news platforms as a forum for public discourse in the context of rapidly shifting political dynamics~\cite{jin2025south}. Most prior research on offensive language in Korean has relied on outdated datasets, with the latest collected in early 2022, limiting their ability to capture recent political developments and emerging patterns of social conflict~\cite{park-etal-2024-predict, park-etal-2023-k, park-etal-2023-feel, jeong-etal-2022-kold, lee-etal-2022-k, kang2022korean}. To advance the field, we constructed a new dataset by curating political news articles and user comments posted throughout 2024 on the most widely used news platform, ensuring it reflects the evolving sociopolitical landscape~\cite{earle2022news, kleinnijenhuis2019combined}.

The news articles are categorized into six topics: \textit{Presidential Office}, \textit{National Assembly / Political Parties}, \textit{North Korea}, \textit{Administration}, \textit{National Defense / Foreign Affairs}, and \textit{General Politics}. Through the fine-grained framework, we constructed the \textbf{\textit{PoliticalK.O}}{\Large\texttwemoji{boxing glove}} dataset (\textit{Political} comments for \textit{K}orean \textit{O}ffensive language), comprising \textit{114,000} articles and \textit{9.28 million} user comments. Since the collected comments contained no ground truth for offensiveness, we employed a diverse set of established offensive language detection methods within a carefully designed framework to assign predicted labels to each comment.

\begin{figure*}[t!]
    \centerline{\includegraphics[width=\textwidth]{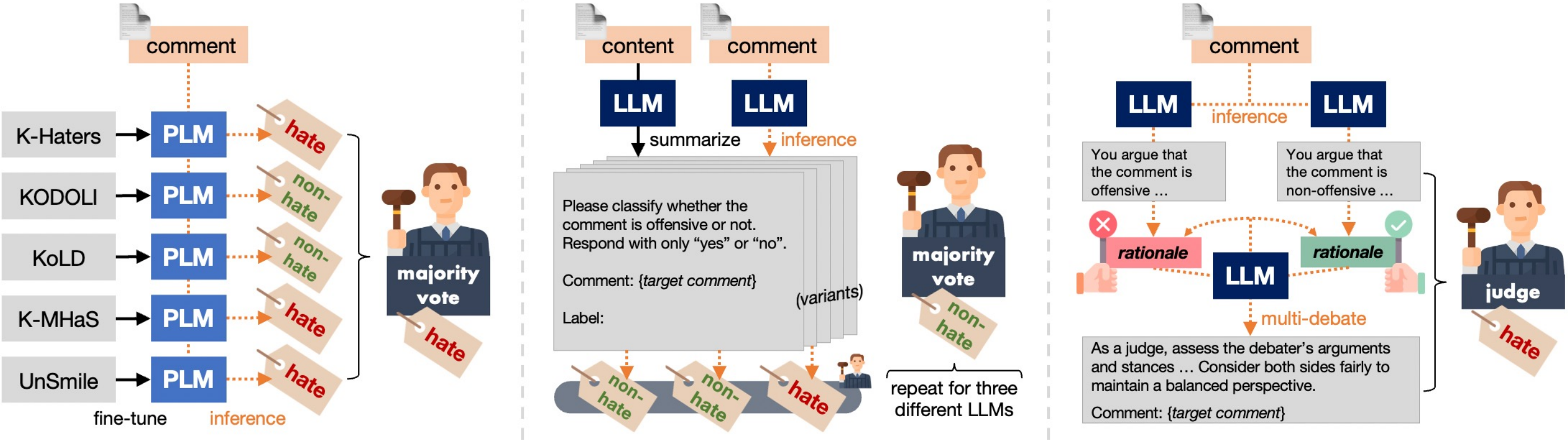}}
    \caption{We adopt three distinct judgments for offensive language detection: \textbf{supervised ensemble judgment} (\texttt{SEJ}) on the left, \textbf{prompt-variants ensemble judgments} (\texttt{PEJ}) in the center, and \textbf{multi-debate reasoning judgment} (\texttt{MRJ}) on the right. Each constitutes a distinct approach to label inference, characterized by its methodology for decision guidance and aggregation, tailored to the newly collected comments from the \textbf{\textit{PoliticalK.O}}{\Large\texttwemoji{boxing glove}}.}
    \label{figure_judgments} 
\end{figure*}

We introduce three main judgments\footnote{Conventional classification tasks evaluate how well predictions align with ground truth, but in our case, we refer to the outcomes of each method as \textit{judgment} due to the absence of ground truth for the unseen comments.}: \textbf{supervised ensemble judgment} (\texttt{SEJ}), \textbf{prompt-variants ensemble judgment} (\texttt{PEJ}), and \textbf{multi-debate reasoning judgment} (\texttt{MRJ}), which are outlined in Figure~\ref{figure_judgments}. First, \texttt{SEJ} utilizes five existing offensive language datasets to fine-tune PLMs to an optimal configuration, combining their predictions through majority voting. Despite the datasets being outdated, we intended to maximize the utilization of meticulously curated data constructed through human annotation~\cite{pandey2022modeling, kwon2022improving}. Following this, \texttt{PEJ} employs three LLMs with the five prompt variants, leveraging contextual information such as explicit definitions of offensive language or article summaries, which are then aggregated by majority voting. We aimed to fully harness the relevant information from an in-context learning perspective~\cite{sun-etal-2023-text, dong2022survey, brown2020language}. Lastly, \texttt{MRJ} assumes that each comment can be interpreted as offensive or non-offensive depending on the perspective, and determines the final label through a multi-agent debate. We intended to enhance the LLM's decision-making capabilities by leveraging contrastive stances~\cite{park-etal-2024-predict, du2023improving}.

We finally derived offensive labels for the unseen comments across three judgments with optimized configurations. Although numerous studies have explored offensive language detection, their applicability to emerging data remains \textit{underexplored}. To fill this gap, we conducted a large-scale analysis on the \textbf{\textit{PoliticalK.O}}{\Large\texttwemoji{boxing glove}} by implementing judgments tailored to the current sociolinguistic landscape. This framework enabled us to examine offensive tendencies of each judgment, identify potential shared decision criteria, and assess label consistency when certain judgments were excluded. Through this analysis, we obtained a granular understanding of how offensive language is perceived in unseen comments from diverse perspectives.

Furthermore, in the absence of ground truth labels for the newly collected comments, we evaluated performance of each judgment through pseudo-labeling~\cite{ahmed2024towards, yang-etal-2023-prototype, zou-caragea-2023-jointmatch}. By treating the aggregated judgments as a ground trust\textsuperscript{\dag}, we conducted an analytic evaluation of the three judgments and their respective components. We then examined how combinations of prior datasets and prompt variants yielded reliable offensive language detection results by leveraging proxy ground trust\textsuperscript{\dag} even without human annotations. Our analysis revealed practical scenarios where a single prompt achieves comparable performance to more resource-intensive methods, highlighting practical approaches for real-world use cases. To facilitate future research in this aspect, we release our dataset and the labels generated by each judgment as open-source resources\footnote{\url{https://github.com/seungukyu/PoliticalK.O}}.

\section{Related Work}
\subsection{Recent Advances in\protect\linebreak\hspace*{\parindent}Offensive Language Detection}

BERT brought significant advances to offensive language detection by enabling bidirectional context modeling~\cite{althobaiti2022bert, roy2022hate, caselli-etal-2021-hatebert}. This capability allowed models to capture context-sensitive features such as indirect hostility and metaphorical expressions that earlier machine learning approaches struggled to address~\cite{ramos2024comprehensive, xiao2024chinese}. Concurrently, researchers began developing more fine-grained and context-aware datasets to account for sociocultural factors~\cite{din2025automatic, pachinger-etal-2024-austrotox, deng-etal-2022-cold, mathew2021hatexplain, rosenthal-etal-2021-solid, zampieri-etal-2019-predicting}. These efforts introduced richer taxonomies and labeling schemes tailored to specific communities.

Since the emergence of LLMs, research has adopted prompt-based approaches such as providing models with explicit criteria for identifying offensiveness~\cite{lu2025unveiling, nghiem-daume-iii-2024-hatecot}, or employing chain-of-thought to guide the reasoning process~\cite{nghiem-daume-iii-2024-hatecot, yang-etal-2023-hare, huang2023chain}. In parallel, a multi-agent method has been proposed to simulate diverse perspectives within decision-making~\cite{park-etal-2024-predict}. This shift reframes the task as capturing a spectrum of interpretations, rather than converging on a single answer.

However, even recent studies in Korean still rely on outdated datasets\footnote{We provided the collection dates of the prior Korean offensive language datasets used in recent studies in Table~\ref{table_prior_datasets_statistics} of Appendix~\ref{appendix_b1}, with the most recent from March 2022, which is now considerably outdated.}, which raises concerns about whether current methods can effectively generalize to emerging sociopolitical discourses and shifting patterns of language use. Several studies have reported that LLMs often struggle to interpret subtler forms of expression, such as political satire and irony~\cite{bojic2025comparing, yi2025irony}. These limitations present challenges to the reliability and societal applicability of offensive language detection systems, emphasizing the need for contemporary datasets and evaluation frameworks that account for ongoing temporal shifts.

Building upon this background, we construct a new dataset focused on contemporary political discourse and conduct a comprehensive evaluation of existing offensive language detection methods to assess their reliability in real-world prediction scenarios. Our quantitative analysis moves beyond conventional ground truth-based accuracy assessments by examining the practical applicability and decision-making consistency of differing judgments, particularly in response to emerging and previously unseen expressions.

\subsection{Sociopolitical Dimensions in Contemporary NLP Tasks}

Prior research in sociopolitical NLP has investigated issues of bias, fairness, and the societal implications of language technologies, in both language models and their downstream applications. These studies have explored challenges such as developing models that account for personal, cultural, and contextual variation~\cite{nguyen2021learning, flek2020returning}. A growing line of work has emphasized the need to examine political polarization and sociodemographic or media-driven bias~\cite{narayanan2023towards, nemeth2023scoping, mohla2023socio}.

Recent advances in LLMs have enabled complex tasks such as public opinion tracking, yet concerns about political bias remain. Numerous studies have shown that LLMs can produce ideologically inconsistent outputs~\cite{aldahoul2025large, potter-etal-2024-hidden, motoki2024more, thapa-etal-2024-side}, potentially influenced by prompt engineering or fine-tuning on politically aligned data~\cite{rozado2024political, bernardelle2024mapping}. Instruction-tuned models in particular have exhibited ideological leanings, raising concerns about their neutrality~\cite{faulborn2025only}, especially as embedded political stances may shift over time~\cite{liu2025turning}. Acknowledging such political bias in datasets and models, we rigorously compare judgments to identify those yielding the most stable and reliable offensiveness detection on large-scale, previously unseen comments.

\subsection{Contrasting Subjectivity in the Interpretation of Offensive Texts}

Research on offensive language detection largely relies on supervised datasets with predefined labels~\cite{korre-etal-2025-untangling, nghiem-etal-2024-define, davidson2017automated, schmidt-wiegand-2017-survey}. These evaluation frameworks emphasize alignment with static annotations that reflect sociocultural norms of a given period~\cite{zhou-etal-2023-cobra}. However, interpretations of offensive content in political discourse are highly context-dependent and vary considerably depending on users' backgrounds and values~\cite{pujari-etal-2024-demand, giorgi2024human}.

A recent study improved robustness in offensive language detection by incorporating annotator disagreement signals, particularly when handling ambiguous or controversial content~\cite{lu2025unveiling}. Nevertheless, evaluations based solely on pre-constructed datasets often overlook the evolving nature of offensive language~\cite{xiao-etal-2024-toxicloakcn, sainz-etal-2023-nlp}. The inherent ambiguity and subjectivity of language complicate annotation consistency among human raters~\cite{rodriguez2024federated, deng-etal-2023-annotate, abercrombie2023consistency}. These challenges are amplified by the emergence of unseen comments, further complicating the identification of robust detection methods. In this context, we construct a topic-diverse dataset of recent political discourse and conduct a comprehensive analysis of diverse judgments to identify strategies best suited for real-world deployment.

\section{Method}
\subsection{Dataset Construction}
\label{section_3.1}

To capture the dynamics of recent political discourse, we constructed \textbf{\textit{PoliticalK.O}}{\Large\texttwemoji{boxing glove}} from Naver\footnote{\url{https://news.naver.com/section/100}}, South Korea's largest news platform, covering all political news articles published in 2024. The dataset includes \textit{114,000} articles and \textit{9.28 million} user comments, along with article summaries and comment threads. The detailed statistics of the dataset are provided in Appendix~\ref{appendix_a}.

\subsection{Supervised Ensemble Judgment}
\label{section_3.2}

We fine-tuned PLMs on five Korean offensive language datasets: K-Haters~\cite{park-etal-2023-k}, KODOLI~\cite{park-etal-2023-feel}, KoLD~\cite{jeong-etal-2022-kold}, K-MHaS~\cite{lee-etal-2022-k}, and UnSmile~\cite{kang2022korean}. Since the original datasets exhibited label imbalance, we re-split them to ensure balanced distributions to prevent potential bias during inference~\cite{shi2022improving}.

We employed multilingual RoBERTa~\cite{DBLP:journals/corr/abs-1911-02116} and KcBERT~\cite{lee2020kcbert} as backbone models and fine-tuned each on five datasets under optimized conditions, resulting in five independently trained models\footnote{The fine-tuning setup and results on each test set are provided in Appendix~\ref{appendix_b1}.}. Although predictions on unseen comments varied depending on the training data, we aggregated them using majority voting. The overall procedure of \texttt{SEJ} is outlined in Algorithm~\ref{algorithm_sej}. While some datasets were outdated, we designed our setup to maximally utilize the strengths of these datasets, leveraging the reliability ensured by their human-curated annotations.

\subsection{Prompt-variants Ensemble Judgment}
\label{section_3.3}

We selected three LLMs recognized for strong performance in Korean—Exaone~\cite{exaone-3.5}, Trillion~\cite{han2025trillion7btechnicalreport}, and HyperclovaX~\cite{yoo2024hyperclova}. Prompt-based methods have gained attention for enabling label inference on unseen data without annotated supervision, in contrast to fine-tuning approaches~\cite{udawatta2024use}. Given that model outputs may vary with prompt formulation even under identical input and model configuration~\cite{voronov-etal-2024-mind}, we employed five prompt variants: \textit{Vanilla (V)}, \textit{Defn (D)}, \textit{Summ (S)}, \textit{FewShots (F)}, and \textit{D+S+F}.

\begin{algorithm}[t!]
\small
\caption{\textbf{Supervised Ensemble Judgment} (\texttt{SEJ})}
\begin{algorithmic}[1]
\State $\{\text{OLD}_{1}, \dots, \text{OLD}_{5}\}$ $\leftarrow$ offensive language datasets
\State model\_pool $\leftarrow \left[ \right]$

\For{i, dataset $\in$ enumerate$\{\text{OLD}_{1}, ..., \text{OLD}_{5}\}$}
    
    \For{$lr \in$ \{1e-5, 2e-5, 3e-5\}}\;\;\textcolor{blue}{\#\# fine-tune}
        \State Train PLMs on train set with $lr$ for 5 epochs
        \If {(best valid loss) or (last) epoch}
            \State model\_pool.append($\text{PLM}_{\textit{trained}}$)
        \EndIf
    \EndFor
    
    \For{$\text{PLM}_{\textit{trained}} \in$ model\_pool}\;\;\textcolor{blue}{\#\# test}
        \State Evaluate on test set
    \EndFor
    \State $\text{best\_PLM}_i \leftarrow \text{PLM}_{\textit{trained}}$ with highest F1 score
\EndFor

\For{comment $\in$ \textbf{\textit{PoliticalK.O}}{\Large\texttwemoji{boxing glove}}}\;\;\textcolor{blue}{\#\# inference}
    \For{$i = 1$ to $5$}
        \State predict$_i$ $\leftarrow$ best\_PLM$_i$(comment)
    \EndFor
    \State label $\leftarrow$ vote(predict$_1$, ..., predict$_5$)
    
\EndFor
\end{algorithmic}
\label{algorithm_sej}
\end{algorithm}

\begin{algorithm}[t!]
\small
\caption{\textbf{Prompt-variants Ensemble Judgment} (\texttt{PEJ})}
\begin{algorithmic}[1]
\State OL $\leftarrow$ refers to offensive language
\For{comment $\in$ \textbf{\textit{PoliticalK.O}}{\Large\texttwemoji{boxing glove}}}

    \State \textcolor{blue}{\#\# prompt construction}
    \State prompt$_{\textit{V}} \leftarrow$ (\textit{basic instruction} for OL, comment)
    \State prompt$_{\textit{D}} \leftarrow$ prompt$_{\textit{V}}$ + \textit{refined definition} for OL
    \State prompt$_{\textit{S}} \leftarrow$ prompt$_{\textit{V}}$ + \textit{summarized source article}
    \State prompt$_{\textit{F}} \leftarrow$ prompt$_{\textit{V}}$ + \textit{few-shot samples}
    \State prompt$_{\textit{D+S+F}} \leftarrow$ combines above three prompts
    
    \State \textcolor{blue}{\#\# inference}
    \For{\textit{ptype} $\in$ $\{\textit{V}, \textit{D}, \textit{S}, \textit{F}, \textit{D+S+F}\}$}\
        \State predict$_1 \leftarrow$ Inference(LLM$_1$, prompt$_{\textit{ptype}}$)
        \State predict$_2 \leftarrow$ Inference(LLM$_2$, prompt$_{\textit{ptype}}$)
        \State predict$_3 \leftarrow$ Inference(LLM$_3$, prompt$_{\textit{ptype}}$)
        \State label$_{\textit{ptype}} \leftarrow$ vote(predict$_1$, ..., predict$_3$)
    \EndFor
    
    \State label $\leftarrow$ vote(label$_{\textit{ptype}}$
    \State \;\;\;\;\;\;\;\;\;\;\;\;\;\;\;\;\;\;\; for \textit{ptype} in \{\textit{V}, \textit{D}, \textit{S}, \textit{F}, \textit{D+S+F}\})
    
\EndFor
\end{algorithmic}
\label{algorithm_pej}
\end{algorithm}

The \textit{Vanilla} provides the comment to be classified, and this standard formulation commonly employed in recent offensive language detection studies~\cite{jaremko-etal-2025-revisiting, pan2024comparing}. The \textit{Defn} provides an explicit definition of offensive language to clarify the model's decision criteria~\cite{lu2025unveiling, nghiem-daume-iii-2024-hatecot}. Given that Korean political discourse frequently references specific politicians or parties, increasing the complexity of assessing offensiveness~\cite{lee-etal-2023-hate}, we refined prior definitions to better reflect these political nuances.

The \textit{Summ} provides background context from the news article. We summarized the source article's title and content into three sentences and appended them, offering contextual grounding for the model~\cite{parvez-2025-chain}. The \textit{FewShots} includes labeled samples from other articles on the same topic as the target comment~\cite{ahmadnia-etal-2025-active, brown2020language}. If the target comment pertains to the topic \textit{North Korea}, few-shot samples were drawn from other articles on that topic. Finally, the \textit{D+S+F} combines all the above elements into a single formulation, supporting more informed predictions through enriched contextual input.

We then applied majority voting to the outputs. The overall procedure of \texttt{PEJ} is outlined in Algorithm~\ref{algorithm_pej}. Although prompt-based inference can depend on individual model behavior, we combined outputs from multiple prompt variants and models to achieve more generalized and robust predictions.

\subsection{Multi-debate Reasoning Judgment}
\label{section3_4}

We extended prior research employing a multi-agent framework for offensive language detection~\cite{park-etal-2024-predict} by refining to better align with judgments on political comments. We evaluated the LLMs in \S Algorithm~\ref{algorithm_pej} using the five datasets from \S Algorithm~\ref{algorithm_sej}, and conducted experiments based on the best-performing model\footnote{The performance of the three LLMs on each of the prior datasets is reported in Appendix~\ref{appendix_b3}. While the main analysis focuses on a single model Trillion, pilot results from the other models are also included in Appendix~\ref{appendix_c2}.}.

We assigned distinct personas to the model, enabling it to interpret comments from different perspectives~\cite{jiang-etal-2024-personallm, hattab2024persona}. Each agent was designed to classify comments as either offensive or non-offensive, generating rationales that illustrate how perspective shapes interpretation. At this stage, we also provided a summary of the source article for each comment to facilitate context-aware assessments~\cite{parvez-2025-chain}.

Subsequently, we instructed agents with opposing viewpoints to generate a stance for each rationale~\cite{hu-etal-2025-debate}. An agent adopting an offensive perspective was asked to debate a rationale from a non-offensive standpoint, and vice versa. Based on all rationales and stances, we employed a judge agent to make the representative label. The overall procedure of \texttt{MRJ} is outlined in Algorithm~\ref{algorithm_mrj}. Through this reasoning process, we aimed to enable agents to analyze offensiveness across a broader spectrum of contextual dimensions.

\begin{algorithm}[t!]
\small
\caption{\textbf{Multi-debate Reasoning Judgment} (\texttt{MRJ})}
\begin{algorithmic}[1]
\State \textcolor{blue}{\#\# model selection}
\State $\{\text{OLD}_{1}, \dots, \text{OLD}_{5}\}$ $\leftarrow$ offensive language datasets

\For{i, dataset $\in$ enumerate$\{\text{OLD}_{1}, ..., \text{OLD}_{5}\}$}
    \State Evaluate LLMs on test set 
\EndFor
\State $\text{best\_LLM} \leftarrow$ LLM with highest F1 score across OLDs

\State \textcolor{blue}{\#\# persona alignment}
\State LLM$_{\textit{O}} \leftarrow$ Initialize $\text{best\_LLM}$ to discuss \textit{offensive}
\State LLM$_{\textit{N}} \leftarrow$ Initialize $\text{best\_LLM}$ to discuss \textit{non-offensive}
\State LLM$_{\textit{Judge}} \leftarrow$ Initialize $\text{best\_LLM}$ to make final decision

\For{comment $\in$ \textbf{\textit{PoliticalK.O}}{\Large\texttwemoji{boxing glove}}}

    \State \textcolor{blue}{\#\# rationale generation}
    \State \textit{summary} $\leftarrow$ \textit{summarized source article}\\
    \;\;\;\;\;\;\;\;\;\;\;\;\;\;\;\;\;\;\;\;\;\;\;\;from \S Algorithm~\ref{algorithm_pej}
    \State rationale$_{\textit{O}} \leftarrow$ Argument(LLM$_{\textit{O}}$, \textit{summary}, comment)
    \State rationale$_{\textit{N}} \leftarrow$ Argument(LLM$_{\textit{N}}$, \textit{summary}, comment)
    
    \State \textcolor{blue}{\#\# discuss on each side}
    \State stance$_{\textit{O}} \leftarrow$ Debate(LLM$_{\textit{O}}$, comment, rationale$_{\textit{N}}$)
    \State stance$_{\textit{N}} \leftarrow$ Debate(LLM$_{\textit{N}}$, comment, rationale$_{\textit{O}}$)

    \State \textcolor{blue}{\#\# final judgment}
    \State label $\leftarrow$ Instruct LLM$_{\textit{Judge}}$ based on\\
    \;\;\;\;\;\;\;\;\;\;\;\;\;\;\;\;\;(rationale$_{\textit{O}}$, rationale$_{\textit{N}}$, stance$_{\textit{O}}$, stance$_{\textit{N}}$)
\EndFor
\end{algorithmic}
\label{algorithm_mrj}
\end{algorithm}

\section{Exploratory Analysis of\protect\linebreak\hspace*{\parindent}Offensiveness across Judgments}
\label{section_4}

We employed the three judgment methods—\texttt{SEJ}, \texttt{PEJ}, and \texttt{MRJ}—on the entire set of comments in the \textbf{\textit{PoliticalK.O}}{\Large\texttwemoji{boxing glove}} to obtain corresponding labels. To examine detection tendencies and potential biases, we first compared the label distributions generated by each judgment. We then calculated the pairwise label overlap ratio to assess the consistency of decision criteria across different judgments.
\begin{align}
\textit{Ratio}_{\textit{overlap}}(A, B) = \frac{1}{n} \sum_{i=1}^{n} \mathbf{1}(A_{i} = B_{i}), \label{eq1}
\end{align}

\begin{table*}[t!]
\small
\centering
\begin{adjustbox}{max width=0.95\textwidth}
\begin{tabular}{l|cccl|cccc}
\hline
\multirow{2}{*}{Topics}                                       & \texttt{SEJ}                      & \texttt{PEJ}                      & \multicolumn{2}{c|}{\texttt{MRJ}}  & \begin{tabular}[c]{@{}c@{}}\texttt{SEJ}\\ $\rightleftharpoons$ \texttt{PEJ}\end{tabular} & \begin{tabular}[c]{@{}c@{}}\texttt{PEJ}\\ $\rightleftharpoons$ \texttt{MRJ}\end{tabular} & \multicolumn{1}{c|}{\begin{tabular}[c]{@{}c@{}}\texttt{MRJ}\\ $\rightleftharpoons$ \texttt{SEJ}\end{tabular}} & \begin{tabular}[c]{@{}c@{}}All\\ Judgments\end{tabular} \\ \cline{2-9} 
                                                              & \multicolumn{4}{c|}{\textit{Label Distribution}}                                                           & \multicolumn{4}{c}{\textit{Pairwise Label Overlap Ratio}}                                                                                                                                                                                                                                                                                                \\ \hline
\textit{Presidential Office}                                  & 59.55 : 40.45                     & 79.69 : 20.31                     & \multicolumn{2}{c|}{80.31 : 19.69} & 75.94                                                                                    & 83.59                                                                                       & \multicolumn{1}{c|}{70.78}                                                                                       & 65.16                                                      \\
\rowcolor[HTML]{EFEFEF} \textit{National Assembly / Political Parties} & 62.58 : 37.42                     & 81.95 : 18.05                     & \multicolumn{2}{c|}{83.85 : 16.15}            & 77.83                                                                                    & 85.89                                                                                       & \multicolumn{1}{c|}{72.22}                                                                                       & 67.97                                                      \\
\textit{North Korea}                                          & 58.12 : 41.88                     & 81.88 : 18.12                     & \multicolumn{2}{c|}{80.35 : 19.65} & 73.77                                                                                    & 84.00                                                                                    & \multicolumn{1}{c|}{69.51}                                                                                    & 63.64                                                      \\
\rowcolor[HTML]{EFEFEF} \textit{Administration}               & 56.56 : 43.44                     & 81.60 : 18.40                     & \multicolumn{2}{c|}{80.03 : 19.97} & 72.92                                                                                    & 84.13                                                                                    & \multicolumn{1}{c|}{69.04}                                                                                    & 63.05                                                      \\
\textit{National Defense / Foreign Affairs}                   & 55.01 : 44.99                     & 74.55 : 25.45                     & \multicolumn{2}{c|}{78.68 : 21.31} & 76.31                                                                                    & 81.35                                                                                    & \multicolumn{1}{c|}{68.43}                                                                                    & 63.05                                                      \\
\rowcolor[HTML]{EFEFEF} \textit{General Politics}             & 62.20 : 37.80                     & 84.99 : 15.01                     & \multicolumn{2}{c|}{82.56 : 17.44}            & 75.40                                                                                    & 86.29                                                                                       & \multicolumn{1}{c|}{72.09}                                                                                       & 66.89                                                      \\ \hline
Total                                                         & \textbf{61.88} : 38.12 & \textbf{83.46} : 16.54 & \multicolumn{2}{c|}{\textbf{82.58} : 17.42}            & 76.09                                                                                    & 85.81                                                                                       & \multicolumn{1}{c|}{71.91}                                                                                       & 66.91                                                      \\ \hline
\end{tabular}
\end{adjustbox}
\caption{Label distribution and pairwise label overlap ratio by topic in the \textbf{\textit{PoliticalK.O}}{\Large\texttwemoji{boxing glove}}. The label distribution values (left : right) represent the proportion of (\textbf{offensive} : non-offensive) labels. We carefully designed each judgment with refined settings to ensure applicability to newly collected political comments.}
\label{table_ld_and_lo}
\end{table*}

The label distributions and pairwise label overlap ratios of judgments across topics are presented in Table~\ref{table_ld_and_lo}. We observed consistently high proportions of offensiveness for all topics, regardless of the judgment type. This trend aligns with longstanding patterns of verbal aggression historically observed in responses to political discourse~\cite{tsoumou2021examination, humprecht2020hostile}. A topic-wise analysis revealed that \textit{National Defense / Foreign Affairs} exhibited a relatively low proportion of offensive comments, whereas \textit{General Politics}, which encompasses a broader range of politician-related issues, showed higher levels. This observation suggests that topic-specific patterns of offensiveness remain consistent across different judgments.

Notably, \texttt{SEJ} inferred from prior datasets yielded a comparatively lower rate of offensive labels with values ranging from approximately 50\% to 60\%. Although the models were trained on explicit examples of offensive language, their reliance on outdated data raises concerns regarding their ability to generalize to contemporary discourse. While it is plausible that LLM-based judgments \texttt{PEJ} and \texttt{MRJ} may overestimate offensiveness, we observed the consistently high label distributions between 70\% and 80\% across a range of prompt variants and multi-debate settings. These results suggest that such outputs are unlikely to result solely from over-detection of previously unseen comments.

We also found that \texttt{PEJ} and \texttt{MRJ} exhibited over 80\% pairwise label overlap, suggesting a notable degree of alignment in decision criteria informed by shared contextual cues and an interconnected reasoning process. In contrast, the overlap ratio decreased when \texttt{SEJ} was included, indicating that its decision boundaries diverge from those of LLM-based judgments, which demonstrate more adaptive and nuanced predictive behavior. When aggregating the judgments across all topics, the overall overlap ratio was approximately 66\%, reflecting a moderate level of consistency.

To assess the impact of individual judgments, we employed a leave-one-out strategy~\cite{inbook, elisseeff2003leave}. We first calculated the agreement score based on the complete set of judgments, then iteratively excluded each judgment to evaluate its effect on the overall score. An increase in agreement upon exclusion ($\Delta _{-k} < 0$) indicates that the removed judgment was misaligned with the consensus. Through this analysis, we aimed to identify whether the aggregated judgment was disproportionately influenced by specific components and to determine more reliable judgment for collective assessments.

From a total of $M$ comments based on $N$ judgments, each label was defined as $x_{i,j}$. The agreement score $f$ was computed using Krippendorff's $\alpha$\footnote{We scaled the obtained scores by multiplying them by 100 to enable more intuitive comparison in our study.}~\cite{krippendorff2011computing}. Let the agreement difference be denoted as $\Delta _{-k}$ obtained by excluding the $k$-th judgment, these are computed as follows:
\begin{align}
\textit{score}_{\textit{total}} &= f\Big(\{\textit{Judgment}_i\}_{i=1}^{N}, \{x_{i, j}\}_{j=1}^{M}\Big), \label{eq2} \\
\textit{score}_{-k} &= f\Big(\{\textit{Judgment}_i\}_{i \neq k}^{N}, \{x_{i, j}\}_{j=1}^{M}\Big), \label{eq3} \\
\Delta _{-k} &= \textit{score}_{\textit{total}} - \textit{score}_{-k}, \label{eq4}
\end{align}

\begin{table}[t!]
\small
\centering
\begin{adjustbox}{max width=\columnwidth}
\begin{tabular}{l|c|cc}
\hline
Topics &
  \begin{tabular}[c]{@{}c@{}}\texttt{SEJ}, \\ \texttt{PEJ}, and \texttt{MRJ}\end{tabular} &
  \begin{tabular}[c]{@{}c@{}}Components\\ in \texttt{SEJ}\end{tabular} &
  \begin{tabular}[c]{@{}c@{}}Components\\ in \texttt{PEJ}\end{tabular} \\ \hline
\textit{Presidential Office}                                                             & 40.83 & 27.23 & 59.94 \\
\rowcolor[HTML]{EFEFEF} \begin{tabular}[c]{@{}l@{}}\textit{National Assembly}\\ / \textit{Political Parties}\end{tabular} & 41.26 & 27.52 & 65.04 \\
\textit{North Korea}                                                                     & 37.86 & 32.46 & 63.98 \\
\rowcolor[HTML]{EFEFEF} \textit{Administration}                                                                  & 37.90 & 30.69 & 62.80 \\
\begin{tabular}[c]{@{}l@{}}\textit{National Defense}\\ / \textit{Foreign Affairs}\end{tabular}    & 41.99 & 31.84 & 62.27 \\
\rowcolor[HTML]{EFEFEF} \textit{General Politics}                                                                & 38.47 & 27.51 & 64.27 \\ \hline
Total                                                                           & 39.57 & 27.60 & 63.96 \\ \hline
\end{tabular}
\end{adjustbox}
\caption{Agreement scores of Krippendorff's $\alpha$ for the three main judgments and within each of \texttt{SEJ} and \texttt{PEJ}.}
\label{table_agreement_scores}
\end{table}

\begin{figure*}[!t]
    \centering
    \begin{minipage}[!t]{0.37\textwidth}
        \centering
        \includegraphics[width=\linewidth]{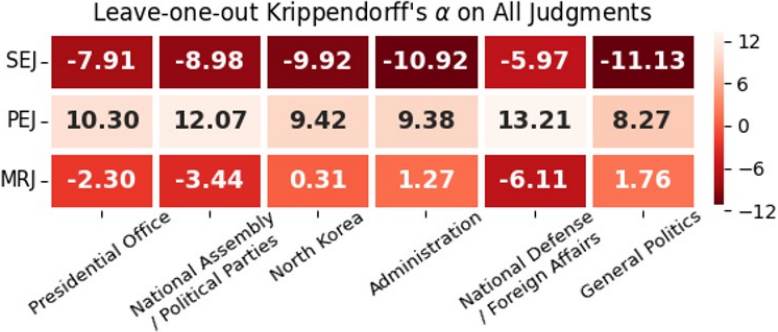}
        \caption{Leave-one-out agreement score differences by excluding each of the main judgment: \texttt{SEJ}, \texttt{PEJ}, and \texttt{MRJ}.}
        \label{figure_loo_all_judgments}
    \end{minipage}
    \hfill
    \begin{minipage}[!t]{0.62\textwidth}
        \centering
        \includegraphics[width=\linewidth]{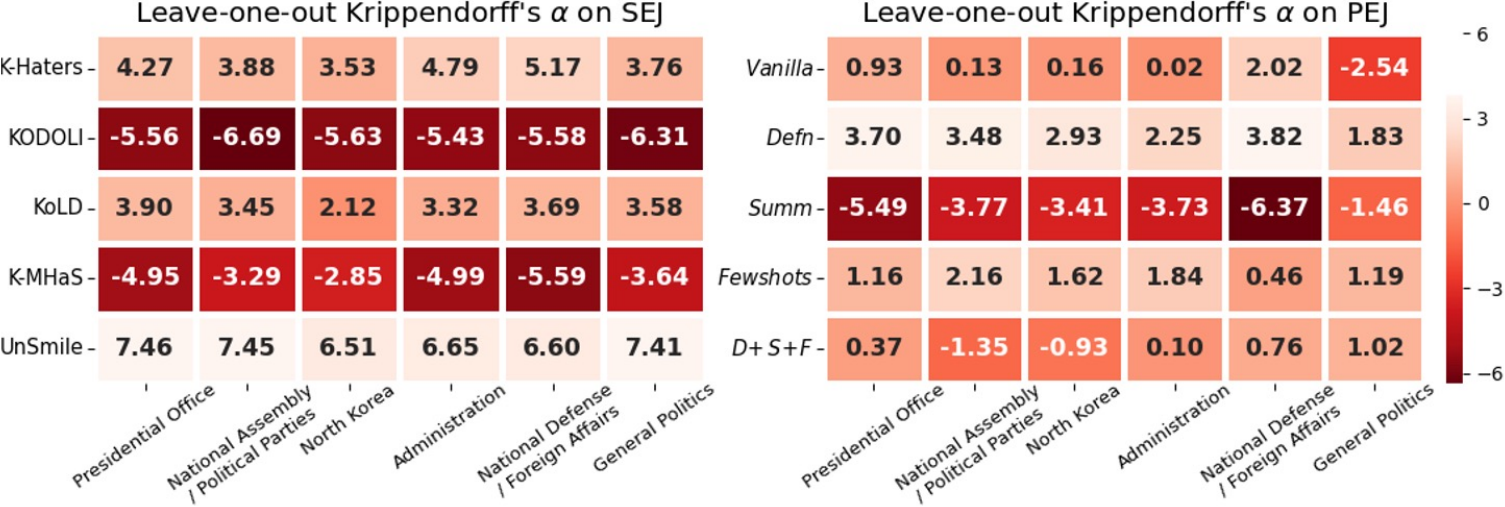}
        \caption{Leave-one-out agreement score differences by excluding each of the component: training dataset or prompt variant, in \texttt{SEJ} and \texttt{PEJ}.}
        \label{figure_loo_sej_pej}
    \end{minipage}
\end{figure*}

The agreement $\textit{score}_{\textit{total}}$ for the three main judgments and within each of \texttt{SEJ} and \texttt{PEJ} are presented in Table~\ref{table_agreement_scores}. We observed that the agreement score across the three main judgments was around 40—a relatively low score considering that each judgment has been widely adopted even in recent studies as a representative approach to offensive language detection. We further investigated how the scores varied depending on the underlying datasets and prompt configurations used in \texttt{SEJ} and \texttt{PEJ}.

One noteworthy aspect is that, despite the large volume of collected comments, especially reaching up to \textit{5.61 million} for the topic \textit{General Politics}, the score of \texttt{PEJ} exceeded 60. This indicates significantly greater consistency compared to \texttt{SEJ}, which remained at around 30. These results suggest that the prompt variants used in our experiment provided more consistent evaluation criteria than those derived from prior datasets, and also imply that prompt-based approaches may ease evaluation and enhance consistency.

The agreement score differences $\Delta _{-k}$ across main and component judgments of \texttt{SEJ} and \texttt{PEJ} are presented in Figure~\ref{figure_loo_all_judgments} and~\ref{figure_loo_sej_pej}. Although it is not possible to conclusively determine which judgment aligns most closely with the ground truth, Figure~\ref{figure_loo_all_judgments} shows that all judgments exhibited substantial discrepancies, with marked shifts upward and downward for \texttt{SEJ} and \texttt{PEJ}. This pattern suggests that relying on pre-existing datasets—even when synthesized from five distinct sources—falls short of the predictive stability demonstrated by more LLM-based approaches. In contrast, \texttt{MRJ} displayed intermediate agreement differences, indicating comparatively greater stability.

We then analyzed each component of \texttt{SEJ} and \texttt{PEJ}, as shown in Figure~\ref{figure_loo_sej_pej}. Larger deviations appeared more often in \texttt{SEJ} than in \texttt{PEJ}, suggesting that the choice of training dataset is a more sensitive factor than prompt design in judgment aggregation. Notably, we found that KODOLI and K-MHaS yielded substantially lower scores, indicating degraded judgment quality. In contrast, UnSmile led to a significant increase, underscoring its importance in aligning inference consistency.

In \texttt{PEJ}, we observed that \textit{Defn} achieved the highest score by providing an explicit definition of offensive language grounded in political context, followed by \textit{FewShots}, which offered topic-relevant samples that enabled more contextual inferences. In contrast, the \textit{Summ} resulted in the lowest score, despite the article summaries exhibiting sufficient factual consistency\footnote{We evaluated the consistency of the summaries in Table~\ref{table_summaries_eval} of Appendix~\ref{appendix_b2}, and found that they generally capture the key content of the corresponding source articles.}. This suggests that the model may have been biased by the article content, potentially leading to under- or overestimation of offensiveness. Moreover, political comments are often driven more by the author's ideological stance rather than the actual content~\cite{han2023news, kubin2021personal}, limiting the effectiveness of article summaries in supporting nuanced judgments in such contexts.

\section{Establishing Ground Trust\textsuperscript{\dag}\protect\linebreak\hspace*{\parindent}for Analytical Evaluation}
In the absence of ground truth labels for offensiveness in our newly constructed dataset, we assessed the performance of each judgment using pseudo-labels~\cite{ahmed2024towards, yang-etal-2023-prototype, zou-caragea-2023-jointmatch}. We treated each carefully designed judgment as a reference point within a trustworthy range, referred to as \textbf{ground trust}\textsuperscript{\dag}. This construct served as a practical proxy for ground truth, enabling systematic evaluation and comparison across offensive language judgments.

\begin{figure}[h!]
    \centerline{\includegraphics[width=\columnwidth]{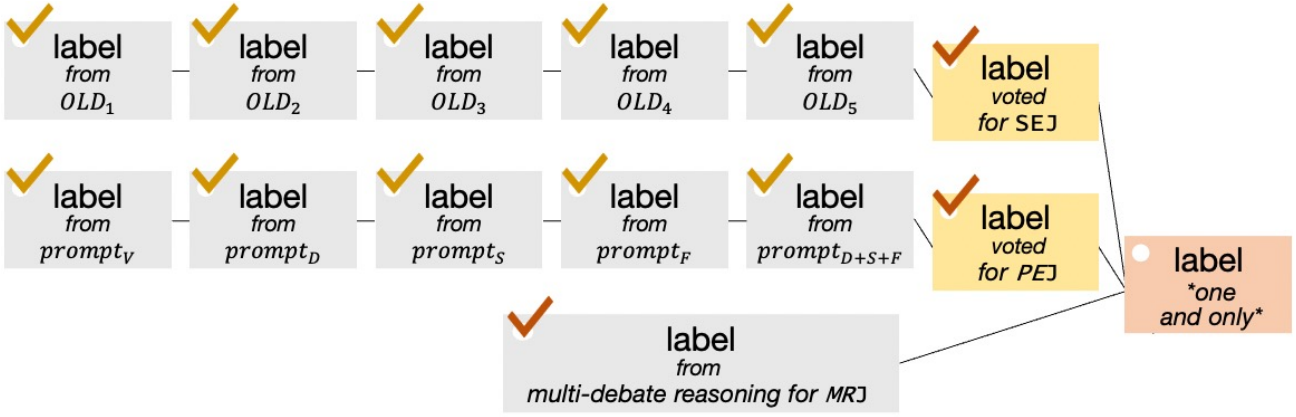}}
    \caption{Hierarchical construction of the ground trust\textsuperscript{\dag} serving as pseudo-labels for evaluating each judgment.}
    \label{figure_ground_trust} 
\end{figure}  

The construction of ground trust\textsuperscript{\dag} from each judgment is illustrated in Figure~\ref{figure_ground_trust}. Individual labels were obtained from \texttt{SEJ}, \texttt{PEJ}, and \texttt{MRJ} (indicated by red check marks), and subsequently aggregated through majority voting (represented by the red box) to form the ground trust\textsuperscript{\dag}. This procedure reflects the integration of multiple optimized conditions to assign a representative label to each comment in the absence of ground truth.

\begin{table*}[t!]
\centering
\small
\begin{adjustbox}{max width=\textwidth}
\begin{tabular}{l|cccc|cccc|cccc}
\hline
\multirow{2}{*}{Topics} & \multicolumn{4}{c|}{\texttt{SEJ}}                 & \multicolumn{4}{c|}{\texttt{PEJ}}                & \multicolumn{4}{c}{\texttt{MRJ}}                 \\ \cline{2-13} 
                  & \multicolumn{1}{c|}{\textit{Acc}} & \textit{P}   & \textit{R}  & \textit{F1} & \multicolumn{1}{c|}{\textit{Acc}} & \textit{P}  & \textit{R}  & \textit{F1} & \multicolumn{1}{c|}{\textit{Acc}} & \textit{P}  & \textit{R}  & \textit{F1} \\ \hline
\textit{Presidential Office} &
  \multicolumn{1}{c|}{81.56} &
  77.52 &
  86.79 &
  78.82 &
  \multicolumn{1}{c|}{94.38} &
  94.31 &
  89.80 &
  91.79 &
  \multicolumn{1}{c|}{89.21} &
  86.67 &
  82.19 &
  84.10 \\
\rowcolor[HTML]{EFEFEF} \textit{National Assembly / Political Parties} &
  \multicolumn{1}{c|}{82.08} &
  76.32 &
  87.83 &
  78.23 &
  \multicolumn{1}{c|}{95.74} &
  95.46 &
  91.26 &
  93.16 &
  \multicolumn{1}{c|}{90.14} &
  87.24 &
  80.94 &
  83.53 \\
\textit{North Korea}       & \multicolumn{1}{c|}{79.64}  & 75.87  & 85.98 & 76.70 & \multicolumn{1}{c|}{94.13}  & 94.95 & 88.09 & 90.95 & \multicolumn{1}{c|}{89.87}  & 86.83 & 83.21 & 84.81 \\
\rowcolor[HTML]{EFEFEF} \textit{Administration}    & \multicolumn{1}{c|}{78.92}  & 75.85  & 85.63 & 76.28 & \multicolumn{1}{c|}{94.00}  & 95.07 & 87.97 & 90.90 & \multicolumn{1}{c|}{90.12}  & 87.60 & 83.71 & 85.41 \\
\textit{National Defense / Foreign Affairs} &
  \multicolumn{1}{c|}{81.70} &
  79.81 &
  86.44 &
  80.28 &
  \multicolumn{1}{c|}{94.61} &
  94.68 &
  91.87 &
  93.14 &
  \multicolumn{1}{c|}{86.73} &
  86.12 &
  79.91 &
  82.20 \\
\rowcolor[HTML]{EFEFEF} \textit{General Politics}  & \multicolumn{1}{c|}{80.60}  & 74.55  & 87.16 & 76.20 & \multicolumn{1}{c|}{94.80}  & 95.60 & 87.45 & 90.84 & \multicolumn{1}{c|}{91.49}  & 87.25 & 84.53 & 85.80 \\ \hline
Total               & \multicolumn{1}{c|}{\textbf{81.09}}  & 75.44 & 87.26 & \textbf{77.11} & \multicolumn{1}{c|}{\textbf{94.99}}  & 95.34 & 88.86 & \textbf{91.65} & \multicolumn{1}{c|}{\textbf{90.82}}  & 87.17 & 83.19 & \textbf{84.96} \\ \hline
\end{tabular}
\end{adjustbox}
\caption{Evaluation results of each judgment based on the constructed ground trust\textsuperscript{\dag} for each topic in the \textbf{\textit{PoliticalK.O}}{\Large\texttwemoji{boxing glove}}. Metrics reported are accuracy, precision, recall, and F1 score. In the absence of ground truth labels, we established ground trust\textsuperscript{\dag} as a pseudo-label within a trustworthy range for evaluation.}
\label{table_ground_trust_eval_3j}
\end{table*}

Although ground trust\textsuperscript{\dag} may not serve as an absolute reference equivalent to conventional ground truth, we regarded it as the most robust pseudo-labeling strategy available in our study, developed through the careful integration of multiple judgments tailored to newly collected comments. Given the ongoing demand for up-to-date dataset construction and the impracticality of exhaustive human annotation, we adopted this automated form of ground trust\textsuperscript{\dag} as a consistent and pragmatic benchmark. The evaluation results for each judgment based on the ground trust\textsuperscript{\dag} are presented in Table~\ref{table_ground_trust_eval_3j}.

We observed that \texttt{PEJ} consistently performed well across all topics, with accuracy and F1 scores exceeding 90, followed by \texttt{MRJ} and \texttt{SEJ}. These results suggest that combining multiple prompt variants proves most effective against our ground trust\textsuperscript{\dag}. Notably, both \texttt{PEJ} and \texttt{MRJ} consistently showed higher precision than recall, reflecting more conservative predictions of offensiveness, yet still outperforming \texttt{SEJ} overall. In contrast, \texttt{SEJ} tended to overestimate offensiveness, as reflected in its consistently higher recall over precision, resulting in lower scores than the other two judgments.

We further evaluated component-level predictions from \texttt{SEJ} and \texttt{PEJ} (indicated by yellow check marks in Figure~\ref{figure_ground_trust}), with results presented in Table~\ref{table_ground_trust_eval_components}. We found that relying on a single dataset, such as KoLD or K-Haters, yielded better performance than combining multiple sources. This highlights the sensitivity of dataset selection in determining the reliability of label inference.

In the case of \texttt{PEJ}, which showed consistent performance across prompt variants, no single case outperformed the aggregated results of all variants. This indicates that while each prompt exhibits robustness on unseen comments, combining outcomes from multiple prompts yields the most reliable performance. Each prompt's performance was comparable to \texttt{MRJ}, suggesting that under constraints on external resources or iterative inference with LLMs, a strategic prompting approach leveraging contextual cues can effectively detect offensiveness in unseen comments\footnote{While our evaluations relied on the ground trust\textsuperscript{\dag}, additional results based on a sampled human-labeled ground truth are also provided in Appendix~\ref{appendix_c1}.}.

\begin{table}[t!]
\small
\centering
\begin{adjustbox}{max width=0.85\columnwidth}
\begin{tabular}{ll|c|ccc}
\hline
\multicolumn{2}{l|}{Each Component} &
  \textit{Acc} &
  \textit{P} &
  \textit{R} &
  \textit{F1} \\ \hline
\multicolumn{1}{l|}{} &
  K-Haters &
  86.13 &
  78.29 &
  82.79 &
  80.12 \\
\multicolumn{1}{l|}{} &
  \cellcolor[HTML]{EFEFEF}KODOLI &
  \cellcolor[HTML]{EFEFEF}42.53 &
  \cellcolor[HTML]{EFEFEF}62.66 &
  \cellcolor[HTML]{EFEFEF}63.76 &
  \cellcolor[HTML]{EFEFEF}42.49 \\
\multicolumn{1}{l|}{} &
  KoLD &
  \textbf{87.39} &
  80.05 &
  83.17 &
  \textbf{81.43} \\
\multicolumn{1}{l|}{} &
  \cellcolor[HTML]{EFEFEF}K-MHaS &
  \cellcolor[HTML]{EFEFEF}45.72 &
  \cellcolor[HTML]{EFEFEF}63.34 &
  \cellcolor[HTML]{EFEFEF}65.83 &
  \cellcolor[HTML]{EFEFEF}45.56 \\
\multicolumn{1}{l|}{\multirow{-5}{*}{\texttt{SEJ}}} &
  UnSmile &
  81.66 &
  74.61 &
  84.25 &
  76.76 \\ \hline
\multicolumn{1}{l|}{} &
  \cellcolor[HTML]{EFEFEF}\textit{Vanilla} &
  \cellcolor[HTML]{EFEFEF}89.54 &
  \cellcolor[HTML]{EFEFEF}82.87 &
  \cellcolor[HTML]{EFEFEF}87.96 &
  \cellcolor[HTML]{EFEFEF}84.98 \\
\multicolumn{1}{l|}{} &
  \textit{Defn} &
  \textbf{92.15} &
  87.21 &
  89.16 &
  \textbf{88.13} \\
\multicolumn{1}{l|}{} &
  \cellcolor[HTML]{EFEFEF}\textit{Summ} &
  \cellcolor[HTML]{EFEFEF}89.86 &
  \cellcolor[HTML]{EFEFEF}90.01 &
  \cellcolor[HTML]{EFEFEF}77.14 &
  \cellcolor[HTML]{EFEFEF}81.45 \\
\multicolumn{1}{l|}{} &
  \textit{FewShots} &
  91.52 &
  89.45 &
  83.03 &
  85.71 \\
\multicolumn{1}{l|}{\multirow{-5}{*}{\texttt{PEJ}}} &
  \cellcolor[HTML]{EFEFEF}\textit{D+S+F} &
  \cellcolor[HTML]{EFEFEF}90.80 &
  \cellcolor[HTML]{EFEFEF}89.83 &
  \cellcolor[HTML]{EFEFEF}80.25 &
  \cellcolor[HTML]{EFEFEF}83.87 \\ \hline
\multicolumn{2}{l|}{\texttt{MRJ}} &
  \textbf{90.82} &
  87.17 &
  83.19 &
  \textbf{84.96} \\ \hline
\end{tabular}
\end{adjustbox}
\caption{Evaluation results of each component from all judgments based on the constructed ground trust\textsuperscript{\dag} for all topics in the \textbf{\textit{PoliticalK.O}}{\Large\texttwemoji{boxing glove}}.}
\label{table_ground_trust_eval_components}
\end{table}

\section{Conclusion}
We constructed the \textbf{\textit{PoliticalK.O}}{\Large\texttwemoji{boxing glove}} dataset—one of the most extensive collections of offensive language with \textit{9.28 million} user comments—to capture the dynamics of contemporary political discourse in Korea. In the absence of ground truth labels, we designed three refined judgments \texttt{SEJ}, \texttt{PEJ}, and \texttt{MRJ} grounded in widely adopted offensive language detection methodologies. Our large-scale analysis revealed notable label distributions and agreement scores on leave-one-out strategy. Notably, when each judgment was evaluated against the ground trust\textsuperscript{\dag}, strategic prompting based on explicit contextual cues consistently achieved performance comparable to more resource-intensive approaches. This result offers a valuable reference point for handling previously unseen comments under real-world constraints, and provides guidance for future scenarios where annotated data or repetitive model access may be limited.

\section*{Limitations}
While our dataset enabled extensive analysis of political comments from 2024, its generalizability to discourse beyond 2025 requires further validation. This study is notable for its exploratory analysis with large-scale collections, in contrast to prior work that heavily relied on outdated datasets. As the analysis is based on Korean, further research is needed to assess its applicability to other languages. We expect the introduced judgments to be flexibly adaptable across languages.

To obtain reliable offensive language judgments on our dataset, we designed a refined experimental framework. Although alternative choices, such as different models and prompt configurations, might further improve the results, we focused on analytical evaluation using established offensive language detection methods that do not require ground truth. We leave the development of a tailored method for complex political discourse for future work.

\section*{Ethics Statement}
In releasing a dataset that contains offensive language, we explicitly state that it must not be used to deliberately target specific individuals or groups. We intended the dataset to support research on the detection and interpretation of offensive language, and this purpose will be clearly defined. Given the nature of online comments, the dataset includes references to political figures and parties. While this aspect aligns with existing offensive language datasets, it raises important questions about how such references affect perceptions of offensiveness, which require further discussion in the context of contemporary discourse.

\section*{Acknowledgments}
This work was supported by the Institute of Information \& Communications Technology Planning \& Evaluation (IITP) grant funded by the Korea government (MSIT) [RS-2021-II211341, Artificial Intelligence Graduate School Program (Chung-Ang University)] and by the National Research Foundation of Korea (NRF) grant funded by the Korea government (MSIT) (RS-2025-00556246).

\bibliography{custom}

\begin{thebibliography}{99}
\providecommand{\natexlab}[1]{#1}

\bibitem[{Abdelsamie et~al.(2024)Abdelsamie, Azab, and Hefny}]{abdelsamie2024comprehensive}
Mahmoud~Mohamed Abdelsamie, Shahira~Shaaban Azab, and Hesham~A Hefny. 2024.
\newblock A comprehensive review on arabic offensive language and hate speech detection on social media: methods, challenges and solutions.
\newblock \emph{Social Network Analysis and Mining}, 14(1):111.

\bibitem[{Abercrombie et~al.(2023)Abercrombie, Rieser, and Hovy}]{abercrombie2023consistency}
Gavin Abercrombie, Verena Rieser, and Dirk Hovy. 2023.
\newblock Consistency is key: Disentangling label variation in natural language processing with intra-annotator agreement.
\newblock \emph{arXiv preprint arXiv:2301.10684}.

\bibitem[{Ahmadnia et~al.(2025)Ahmadnia, Yousefi~Jordehi, Hosseini Khasheh~Heyran, Mirroshandel, Rambow, and Caragea}]{ahmadnia-etal-2025-active}
Saeed Ahmadnia, Arash Yousefi~Jordehi, Mahsa Hosseini Khasheh~Heyran, Seyed~Abolghasem Mirroshandel, Owen Rambow, and Cornelia Caragea. 2025.
\newblock \href {https://aclanthology.org/2025.naacl-long.340/} {Active few-shot learning for text classification}.
\newblock In \emph{Proceedings of the 2025 Conference of the Nations of the Americas Chapter of the Association for Computational Linguistics: Human Language Technologies (Volume 1: Long Papers)}, pages 6677--6694, Albuquerque, New Mexico. Association for Computational Linguistics.

\bibitem[{Ahmed et~al.(2024)Ahmed, Wen, Ao, Pan, Su, Cao, and Liu}]{ahmed2024towards}
Murtadha Ahmed, Bo~Wen, Luo Ao, Shengfeng Pan, Jianlin Su, Xinxin Cao, and Yunfeng Liu. 2024.
\newblock Towards robust learning with noisy and pseudo labels for text classification.
\newblock \emph{Information Sciences}, 661:120160.

\bibitem[{Aklouche et~al.(2024)Aklouche, Bazine, and Ghalia-Bououchma}]{aklouche2024offensive}
Billel Aklouche, Yousra Bazine, and Zoumrouda Ghalia-Bououchma. 2024.
\newblock Offensive language and hate speech detection using transformers and ensemble learning approaches.
\newblock \emph{Computaci{\'o}n y Sistemas}, 28(3):1031--1039.

\bibitem[{Aldahoul et~al.(2025)Aldahoul, Ibrahim, Varvello, Kaufman, Rahwan, and Zaki}]{aldahoul2025large}
Nouar Aldahoul, Hazem Ibrahim, Matteo Varvello, Aaron Kaufman, Talal Rahwan, and Yasir Zaki. 2025.
\newblock Large language models are often politically extreme, usually ideologically inconsistent, and persuasive even in informational contexts.
\newblock \emph{arXiv preprint arXiv:2505.04171}.

\bibitem[{Althobaiti(2022)}]{althobaiti2022bert}
Maha~Jarallah Althobaiti. 2022.
\newblock Bert-based approach to arabic hate speech and offensive language detection in twitter: exploiting emojis and sentiment analysis.
\newblock \emph{International Journal of Advanced Computer Science and Applications}, 13(5).

\bibitem[{Bernardelle et~al.(2024)Bernardelle, Fr{\"o}hling, Civelli, Lunardi, Roitero, and Demartini}]{bernardelle2024mapping}
Pietro Bernardelle, Leon Fr{\"o}hling, Stefano Civelli, Riccardo Lunardi, Kevin Roitero, and Gianluca Demartini. 2024.
\newblock Mapping and influencing the political ideology of large language models using synthetic personas.
\newblock \emph{arXiv preprint arXiv:2412.14843}.

\bibitem[{Boji{\'c} et~al.(2025)Boji{\'c}, Zagovora, Zelenkauskaite, Vukovi{\'c}, {\v{C}}abarkapa, Veseljevi{\'c}~Jerkovi{\'c}, and Jovan{\v{c}}evi{\'c}}]{bojic2025comparing}
Ljubi{\v{s}}a Boji{\'c}, Olga Zagovora, Asta Zelenkauskaite, Vuk Vukovi{\'c}, Milan {\v{C}}abarkapa, Selma Veseljevi{\'c}~Jerkovi{\'c}, and Ana Jovan{\v{c}}evi{\'c}. 2025.
\newblock Comparing large language models and human annotators in latent content analysis of sentiment, political leaning, emotional intensity and sarcasm.
\newblock \emph{Scientific reports}, 15(1):11477.

\bibitem[{Bragg et~al.(2021)Bragg, Cohan, Lo, and Beltagy}]{bragg2021flex}
Jonathan Bragg, Arman Cohan, Kyle Lo, and Iz~Beltagy. 2021.
\newblock Flex: Unifying evaluation for few-shot nlp.
\newblock \emph{Advances in neural information processing systems}, 34:15787--15800.

\bibitem[{Brown et~al.(2020)Brown, Mann, Ryder, Subbiah, Kaplan, Dhariwal, Neelakantan, Shyam, Sastry, Askell et~al.}]{brown2020language}
Tom Brown, Benjamin Mann, Nick Ryder, Melanie Subbiah, Jared~D Kaplan, Prafulla Dhariwal, Arvind Neelakantan, Pranav Shyam, Girish Sastry, Amanda Askell, et~al. 2020.
\newblock Language models are few-shot learners.
\newblock \emph{Advances in neural information processing systems}, 33:1877--1901.

\bibitem[{Caselli et~al.(2021)Caselli, Basile, Mitrovi{\'c}, and Granitzer}]{caselli-etal-2021-hatebert}
Tommaso Caselli, Valerio Basile, Jelena Mitrovi{\'c}, and Michael Granitzer. 2021.
\newblock \href {https://doi.org/10.18653/v1/2021.woah-1.3} {{H}ate{BERT}: Retraining {BERT} for abusive language detection in {E}nglish}.
\newblock In \emph{Proceedings of the 5th Workshop on Online Abuse and Harms (WOAH 2021)}, pages 17--25, Online. Association for Computational Linguistics.

\bibitem[{Chen et~al.(2023)Chen, Sun, Zhang, and Zhang}]{chen2023automatic}
Huiyao Chen, Yueheng Sun, Meishan Zhang, and Min Zhang. 2023.
\newblock Automatic noise generation and reduction for text classification.
\newblock \emph{IEEE/ACM Transactions on Audio, Speech, and Language Processing}, 32:139--150.

\bibitem[{Chhabra et~al.(2024)Chhabra, Askari, and Mohapatra}]{chhabra-etal-2024-revisiting}
Anshuman Chhabra, Hadi Askari, and Prasant Mohapatra. 2024.
\newblock \href {https://doi.org/10.18653/v1/2024.naacl-short.1} {Revisiting zero-shot abstractive summarization in the era of large language models from the perspective of position bias}.
\newblock In \emph{Proceedings of the 2024 Conference of the North American Chapter of the Association for Computational Linguistics: Human Language Technologies (Volume 2: Short Papers)}, pages 1--11, Mexico City, Mexico. Association for Computational Linguistics.

\bibitem[{Conneau et~al.(2019)Conneau, Khandelwal, Goyal, Chaudhary, Wenzek, Guzm{\'{a}}n, Grave, Ott, Zettlemoyer, and Stoyanov}]{DBLP:journals/corr/abs-1911-02116}
Alexis Conneau, Kartikay Khandelwal, Naman Goyal, Vishrav Chaudhary, Guillaume Wenzek, Francisco Guzm{\'{a}}n, Edouard Grave, Myle Ott, Luke Zettlemoyer, and Veselin Stoyanov. 2019.
\newblock \href {https://arxiv.org/abs/1911.02116} {Unsupervised cross-lingual representation learning at scale}.
\newblock \emph{CoRR}, abs/1911.02116.

\bibitem[{Davidson et~al.(2017)Davidson, Warmsley, Macy, and Weber}]{davidson2017automated}
Thomas Davidson, Dana Warmsley, Michael Macy, and Ingmar Weber. 2017.
\newblock Automated hate speech detection and the problem of offensive language.
\newblock In \emph{Proceedings of the international AAAI conference on web and social media}, volume~11, pages 512--515.

\bibitem[{Deng et~al.(2022)Deng, Zhou, Sun, Zheng, Mi, Meng, and Huang}]{deng-etal-2022-cold}
Jiawen Deng, Jingyan Zhou, Hao Sun, Chujie Zheng, Fei Mi, Helen Meng, and Minlie Huang. 2022.
\newblock \href {https://doi.org/10.18653/v1/2022.emnlp-main.796} {{COLD}: A benchmark for {C}hinese offensive language detection}.
\newblock In \emph{Proceedings of the 2022 Conference on Empirical Methods in Natural Language Processing}, pages 11580--11599, Abu Dhabi, United Arab Emirates. Association for Computational Linguistics.

\bibitem[{Deng et~al.(2023)Deng, Zhang, Liu, Wu, Wang, and Mihalcea}]{deng-etal-2023-annotate}
Naihao Deng, Xinliang Zhang, Siyang Liu, Winston Wu, Lu~Wang, and Rada Mihalcea. 2023.
\newblock \href {https://doi.org/10.18653/v1/2023.findings-emnlp.832} {You are what you annotate: Towards better models through annotator representations}.
\newblock In \emph{Findings of the Association for Computational Linguistics: EMNLP 2023}, pages 12475--12498, Singapore. Association for Computational Linguistics.

\bibitem[{Din et~al.(2025)Din, Khusro, Khan, Ahmad, Ali, and Ghazal}]{din2025automatic}
Salah~Ud Din, Shah Khusro, Farman~Ali Khan, Munir Ahmad, Oualid Ali, and Taher~M Ghazal. 2025.
\newblock An automatic approach for the identification of offensive language in perso-arabic urdu language: Dataset creation and evaluation.
\newblock \emph{IEEE Access}.

\bibitem[{Dong et~al.(2022)Dong, Li, Dai, Zheng, Ma, Li, Xia, Xu, Wu, Liu et~al.}]{dong2022survey}
Qingxiu Dong, Lei Li, Damai Dai, Ce~Zheng, Jingyuan Ma, Rui Li, Heming Xia, Jingjing Xu, Zhiyong Wu, Tianyu Liu, et~al. 2022.
\newblock A survey on in-context learning.
\newblock \emph{arXiv preprint arXiv:2301.00234}.

\bibitem[{Du et~al.(2023)Du, Li, Torralba, Tenenbaum, and Mordatch}]{du2023improving}
Yilun Du, Shuang Li, Antonio Torralba, Joshua~B Tenenbaum, and Igor Mordatch. 2023.
\newblock Improving factuality and reasoning in language models through multiagent debate.
\newblock In \emph{Forty-first International Conference on Machine Learning}.

\bibitem[{Earle and Hodson(2022)}]{earle2022news}
Megan Earle and Gordon Hodson. 2022.
\newblock News media impact on sociopolitical attitudes.
\newblock \emph{Plos one}, 17(3):e0264031.

\bibitem[{Elisseeff et~al.(2003)Elisseeff, Pontil et~al.}]{elisseeff2003leave}
Andr{\'e} Elisseeff, Massimiliano Pontil, et~al. 2003.
\newblock Leave-one-out error and stability of learning algorithms with applications.
\newblock \emph{NATO science series sub series iii computer and systems sciences}, 190:111--130.

\bibitem[{Faulborn et~al.(2025)Faulborn, Sen, Pellert, Spitz, and Garcia}]{faulborn2025only}
Mats Faulborn, Indira Sen, Max Pellert, Andreas Spitz, and David Garcia. 2025.
\newblock Only a little to the left: A theory-grounded measure of political bias in large language models.
\newblock \emph{arXiv preprint arXiv:2503.16148}.

\bibitem[{Flek(2020)}]{flek2020returning}
Lucie Flek. 2020.
\newblock Returning the n to nlp: Towards contextually personalized classification models.
\newblock In \emph{Proceedings of the 58th annual meeting of the association for computational linguistics}, pages 7828--7838.

\bibitem[{Gao et~al.(2023)Gao, Ruan, Sun, Yin, Yang, and Wan}]{gao2023human}
Mingqi Gao, Jie Ruan, Renliang Sun, Xunjian Yin, Shiping Yang, and Xiaojun Wan. 2023.
\newblock Human-like summarization evaluation with chatgpt.
\newblock \emph{arXiv preprint arXiv:2304.02554}.

\bibitem[{Ghenai et~al.(2025)Ghenai, Noorian, Moradisani, Abadeh, Erentzen, and Zarrinkalam}]{GHENAI2025104079}
Amira Ghenai, Zeinab Noorian, Hadiseh Moradisani, Parya Abadeh, Caroline Erentzen, and Fattane Zarrinkalam. 2025.
\newblock \href {https://doi.org/10.1016/j.ipm.2025.104079} {Exploring hate speech dynamics: The emotional, linguistic, and thematic impact on social media users}.
\newblock \emph{Information Processing \& Management}, 62(3):104079.

\bibitem[{Giorgi et~al.(2024)Giorgi, Cima, Fagni, Avvenuti, and Cresci}]{giorgi2024human}
Tommaso Giorgi, Lorenzo Cima, Tiziano Fagni, Marco Avvenuti, and Stefano Cresci. 2024.
\newblock Human and llm biases in hate speech annotations: A socio-demographic analysis of annotators and targets.
\newblock \emph{arXiv preprint arXiv:2410.07991}.

\bibitem[{Han et~al.(2023)Han, Lee, Lee, and Cha}]{han2023news}
Jiyoung Han, Youngin Lee, Junbum Lee, and Meeyoung Cha. 2023.
\newblock News comment sections and online echo chambers: The ideological alignment between partisan news stories and their user comments.
\newblock \emph{Journalism}, 24(8):1836--1856.

\bibitem[{Han et~al.(2025)Han, Suk, An, Kim, Kim, Yang, Choi, and Shin}]{han2025trillion7btechnicalreport}
Sungjun Han, Juyoung Suk, Suyeong An, Hyungguk Kim, Kyuseok Kim, Wonsuk Yang, Seungtaek Choi, and Jamin Shin. 2025.
\newblock \href {https://arxiv.org/abs/2504.15431} {Trillion 7b technical report}.
\newblock \emph{Preprint}, arXiv:2504.15431.

\bibitem[{Hattab et~al.(2024)Hattab, An{\v{z}}el, Dubey, Ezekannagha, Yang, and {\.I}lgen}]{hattab2024persona}
Georges Hattab, Aleksandar An{\v{z}}el, Akshat Dubey, Chisom Ezekannagha, Zewen Yang, and Bahar {\.I}lgen. 2024.
\newblock Persona adaptable strategies make large language models tractable.
\newblock In \emph{Proceedings of the 2024 8th International Conference on Natural Language Processing and Information Retrieval}, pages 24--31.

\bibitem[{Houamegni and Gedikli(2025)}]{houamegni2025evaluating}
Lionel Richy~Panlap Houamegni and Fatih Gedikli. 2025.
\newblock Evaluating the effectiveness of large language models in automated news article summarization.
\newblock \emph{arXiv preprint arXiv:2502.17136}.

\bibitem[{Hu et~al.(2025)Hu, Chan, Li, and Yin}]{hu-etal-2025-debate}
Zhe Hu, Hou~Pong Chan, Jing Li, and Yu~Yin. 2025.
\newblock \href {https://aclanthology.org/2025.coling-main.314/} {Debate-to-write: A persona-driven multi-agent framework for diverse argument generation}.
\newblock In \emph{Proceedings of the 31st International Conference on Computational Linguistics}, pages 4689--4703, Abu Dhabi, UAE. Association for Computational Linguistics.

\bibitem[{Huang et~al.(2023)Huang, Kwak, and An}]{huang2023chain}
Fan Huang, Haewoon Kwak, and Jisun An. 2023.
\newblock Chain of explanation: New prompting method to generate quality natural language explanation for implicit hate speech.
\newblock In \emph{Companion proceedings of the ACM Web conference 2023}, pages 90--93.

\bibitem[{Humprecht et~al.(2020)Humprecht, Hellmueller, and Lischka}]{humprecht2020hostile}
Edda Humprecht, Lea Hellmueller, and Juliane~A Lischka. 2020.
\newblock Hostile emotions in news comments: A cross-national analysis of facebook discussions.
\newblock \emph{Social Media+ Society}, 6(1):2056305120912481.

\bibitem[{Jaremko et~al.(2025)Jaremko, Gromann, and Wiegand}]{jaremko-etal-2025-revisiting}
Julia Jaremko, Dagmar Gromann, and Michael Wiegand. 2025.
\newblock \href {https://aclanthology.org/2025.coling-main.262/} {Revisiting implicitly abusive language detection: Evaluating {LLM}s in zero-shot and few-shot settings}.
\newblock In \emph{Proceedings of the 31st International Conference on Computational Linguistics}, pages 3879--3898, Abu Dhabi, UAE. Association for Computational Linguistics.

\bibitem[{Jeong et~al.(2022)Jeong, Oh, Lee, Ahn, Moon, Park, and Oh}]{jeong-etal-2022-kold}
Younghoon Jeong, Juhyun Oh, Jongwon Lee, Jaimeen Ahn, Jihyung Moon, Sungjoon Park, and Alice Oh. 2022.
\newblock \href {https://doi.org/10.18653/v1/2022.emnlp-main.744} {{KOLD}: {K}orean offensive language dataset}.
\newblock In \emph{Proceedings of the 2022 Conference on Empirical Methods in Natural Language Processing}, pages 10818--10833, Abu Dhabi, United Arab Emirates. Association for Computational Linguistics.

\bibitem[{Jia et~al.(2023)Jia, Ren, Liu, and Zhu}]{jia-etal-2023-zero}
Qi~Jia, Siyu Ren, Yizhu Liu, and Kenny Zhu. 2023.
\newblock \href {https://doi.org/10.18653/v1/2023.emnlp-main.679} {Zero-shot faithfulness evaluation for text summarization with foundation language model}.
\newblock In \emph{Proceedings of the 2023 Conference on Empirical Methods in Natural Language Processing}, pages 11017--11031, Singapore. Association for Computational Linguistics.

\bibitem[{Jiang et~al.(2024)Jiang, Zhang, Cao, Breazeal, Roy, and Kabbara}]{jiang-etal-2024-personallm}
Hang Jiang, Xiajie Zhang, Xubo Cao, Cynthia Breazeal, Deb Roy, and Jad Kabbara. 2024.
\newblock \href {https://doi.org/10.18653/v1/2024.findings-naacl.229} {{P}ersona{LLM}: Investigating the ability of large language models to express personality traits}.
\newblock In \emph{Findings of the Association for Computational Linguistics: NAACL 2024}, pages 3605--3627, Mexico City, Mexico. Association for Computational Linguistics.

\bibitem[{Jin(2025)}]{jin2025south}
Youngjae Jin. 2025.
\newblock South korea in 2024: Political uncertainty, economic challenges, and cultural ascendancy.
\newblock \emph{Asian Survey}, 65(2):214--227.

\bibitem[{Kang et~al.(2022)Kang, Kwon, Lee, Nam, Song, and Suh}]{kang2022korean}
TaeYoung Kang, Eunrang Kwon, Junbum Lee, Youngeun Nam, Junmo Song, and JeongKyu Suh. 2022.
\newblock Korean online hate speech dataset for multilabel classification: How can social science improve dataset on hate speech?
\newblock \emph{arXiv preprint arXiv:2204.03262}.

\bibitem[{Kaur et~al.(2024)Kaur, Singh, and Kaushal}]{kaur2024deep}
Simrat Kaur, Sarbjeet Singh, and Sakshi Kaushal. 2024.
\newblock Deep learning-based approaches for abusive content detection and classification for multi-class online user-generated data.
\newblock \emph{International Journal of Cognitive Computing in Engineering}, 5:104--122.

\bibitem[{Kleinnijenhuis et~al.(2019)Kleinnijenhuis, Van~Hoof, and Van~Atteveldt}]{kleinnijenhuis2019combined}
Jan Kleinnijenhuis, Anita~MJ Van~Hoof, and Wouter Van~Atteveldt. 2019.
\newblock The combined effects of mass media and social media on political perceptions and preferences.
\newblock \emph{Journal of Communication}, 69(6):650--673.

\bibitem[{Korre et~al.(2025)Korre, Muti, Ruggeri, and Barr{\'o}n-Cede{\~n}o}]{korre-etal-2025-untangling}
Katerina Korre, Arianna Muti, Federico Ruggeri, and Alberto Barr{\'o}n-Cede{\~n}o. 2025.
\newblock \href {https://aclanthology.org/2025.findings-naacl.175/} {Untangling hate speech definitions: A semantic componential analysis across cultures and domains}.
\newblock In \emph{Findings of the Association for Computational Linguistics: NAACL 2025}, pages 3184--3198, Albuquerque, New Mexico. Association for Computational Linguistics.

\bibitem[{Krippendorff(2011)}]{krippendorff2011computing}
Klaus Krippendorff. 2011.
\newblock Computing krippendorff’s alpha-reliability.

\bibitem[{Kubin et~al.(2021)Kubin, Puryear, Schein, and Gray}]{kubin2021personal}
Emily Kubin, Curtis Puryear, Chelsea Schein, and Kurt Gray. 2021.
\newblock Personal experiences bridge moral and political divides better than facts.
\newblock \emph{Proceedings of the National Academy of Sciences}, 118(6):e2008389118.

\bibitem[{Kwon et~al.(2022)Kwon, Park, Byon, and Lee}]{kwon2022improving}
Eunjung Kwon, Hyunho Park, Sungwon Byon, and Kyu-Chul Lee. 2022.
\newblock Improving text classification performance through data labeling adjustment.
\newblock In \emph{2022 13th International Conference on Information and Communication Technology Convergence (ICTC)}, pages 2277--2279. IEEE.

\bibitem[{Kwon et~al.(2023)Kwon, Li, Zhuang, Sheng, Zheng, Yu, Gonzalez, Zhang, and Stoica}]{kwon2023efficient}
Woosuk Kwon, Zhuohan Li, Siyuan Zhuang, Ying Sheng, Lianmin Zheng, Cody~Hao Yu, Joseph~E. Gonzalez, Hao Zhang, and Ion Stoica. 2023.
\newblock Efficient memory management for large language model serving with pagedattention.
\newblock In \emph{Proceedings of the ACM SIGOPS 29th Symposium on Operating Systems Principles}.

\bibitem[{Lee et~al.(2022)Lee, Lim, Lee, Jo, Kim, Yoon, and Han}]{lee-etal-2022-k}
Jean Lee, Taejun Lim, Heejun Lee, Bogeun Jo, Yangsok Kim, Heegeun Yoon, and Soyeon~Caren Han. 2022.
\newblock \href {https://aclanthology.org/2022.coling-1.311/} {K-{MH}a{S}: A multi-label hate speech detection dataset in {K}orean online news comment}.
\newblock In \emph{Proceedings of the 29th International Conference on Computational Linguistics}, pages 3530--3538, Gyeongju, Republic of Korea. International Committee on Computational Linguistics.

\bibitem[{Lee(2020)}]{lee2020kcbert}
Junbum Lee. 2020.
\newblock Kcbert: Korean comments bert.
\newblock In \emph{Proceedings of the 32nd Annual Conference on Human and Cognitive Language Technology}, pages 437--440.

\bibitem[{Lee et~al.(2023)Lee, Jung, and Oh}]{lee-etal-2023-hate}
Nayeon Lee, Chani Jung, and Alice Oh. 2023.
\newblock \href {https://doi.org/10.18653/v1/2023.c3nlp-1.5} {Hate speech classifiers are culturally insensitive}.
\newblock In \emph{Proceedings of the First Workshop on Cross-Cultural Considerations in NLP (C3NLP)}, pages 35--46, Dubrovnik, Croatia. Association for Computational Linguistics.

\bibitem[{{LG AI Research}(2024)}]{exaone-3.5}
{LG AI Research}. 2024.
\newblock Exaone 3.5: Series of large language models for real-world use cases.
\newblock \emph{arXiv preprint arXiv:https://arxiv.org/abs/2412.04862}.

\bibitem[{Liu et~al.(2025)Liu, Panwang, and Gu}]{liu2025turning}
Yifei Liu, Yuang Panwang, and Chao Gu. 2025.
\newblock “turning right”? an experimental study on the political value shift in large language models.
\newblock \emph{Humanities and Social Sciences Communications}, 12(1):1--10.

\bibitem[{Lu et~al.(2025)Lu, Ma, Wang, Xiao, Lee, Xu, Yang, and Lin}]{lu2025unveiling}
Junyu Lu, Kai Ma, Kaichun Wang, Kelaiti Xiao, Roy Ka-Wei Lee, Bo~Xu, Liang Yang, and Hongfei Lin. 2025.
\newblock Unveiling the capabilities of large language models in detecting offensive language with annotation disagreement.
\newblock \emph{arXiv preprint arXiv:2502.06207}.

\bibitem[{Mathew et~al.(2021)Mathew, Saha, Yimam, Biemann, Goyal, and Mukherjee}]{mathew2021hatexplain}
Binny Mathew, Punyajoy Saha, Seid~Muhie Yimam, Chris Biemann, Pawan Goyal, and Animesh Mukherjee. 2021.
\newblock Hatexplain: A benchmark dataset for explainable hate speech detection.
\newblock In \emph{Proceedings of the AAAI conference on artificial intelligence}, volume~35, pages 14867--14875.

\bibitem[{Mnassri et~al.(2024)Mnassri, Farahbakhsh, Chalehchaleh, Rajapaksha, Jafari, Li, and Crespi}]{mnassri2024survey}
Khouloud Mnassri, Reza Farahbakhsh, Razieh Chalehchaleh, Praboda Rajapaksha, Amir~Reza Jafari, Guanlin Li, and Noel Crespi. 2024.
\newblock A survey on multi-lingual offensive language detection.
\newblock \emph{PeerJ Computer Science}, 10:e1934.

\bibitem[{Mohla and Guha(2023)}]{mohla2023socio}
Satyam Mohla and Anupam Guha. 2023.
\newblock Socio-economic landscape of digital transformation \& public nlp systems: A critical review.
\newblock \emph{arXiv preprint arXiv:2304.01651}.

\bibitem[{Motoki et~al.(2024)Motoki, Pinho~Neto, and Rodrigues}]{motoki2024more}
Fabio Motoki, Valdemar Pinho~Neto, and Victor Rodrigues. 2024.
\newblock More human than human: measuring chatgpt political bias.
\newblock \emph{Public Choice}, 198(1):3--23.

\bibitem[{Narayanan~Venkit(2023)}]{narayanan2023towards}
Pranav Narayanan~Venkit. 2023.
\newblock Towards a holistic approach: Understanding sociodemographic biases in nlp models using an interdisciplinary lens.
\newblock In \emph{Proceedings of the 2023 AAAI/ACM Conference on AI, Ethics, and Society}, pages 1004--1005.

\bibitem[{N{\'e}meth(2023)}]{nemeth2023scoping}
Ren{\'a}ta N{\'e}meth. 2023.
\newblock A scoping review on the use of natural language processing in research on political polarization: trends and research prospects.
\newblock \emph{Journal of computational social science}, 6(1):289--313.

\bibitem[{Nghiem and Daum{\'e}~Iii(2024)}]{nghiem-daume-iii-2024-hatecot}
Huy Nghiem and Hal Daum{\'e}~Iii. 2024.
\newblock \href {https://doi.org/10.18653/v1/2024.findings-emnlp.343} {{H}ate{COT}: An explanation-enhanced dataset for generalizable offensive speech detection via large language models}.
\newblock In \emph{Findings of the Association for Computational Linguistics: EMNLP 2024}, pages 5938--5956, Miami, Florida, USA. Association for Computational Linguistics.

\bibitem[{Nghiem et~al.(2024)Nghiem, Gupta, and Morstatter}]{nghiem-etal-2024-define}
Huy Nghiem, Umang Gupta, and Fred Morstatter. 2024.
\newblock \href {https://aclanthology.org/2024.eacl-long.78/} {{\textquotedblleft}define your terms{\textquotedblright} : Enhancing efficient offensive speech classification with definition}.
\newblock In \emph{Proceedings of the 18th Conference of the European Chapter of the Association for Computational Linguistics (Volume 1: Long Papers)}, pages 1293--1309, St. Julian{'}s, Malta. Association for Computational Linguistics.

\bibitem[{Nguyen et~al.(2021)Nguyen, Rosseel, and Grieve}]{nguyen2021learning}
Dong Nguyen, Laura Rosseel, and Jack Grieve. 2021.
\newblock On learning and representing social meaning in nlp: a sociolinguistic perspective.
\newblock In \emph{Proceedings of the 2021 Conference of the North American Chapter of the Association for Computational Linguistics: Human language technologies}, pages 603--612.

\bibitem[{Pachinger et~al.(2024)Pachinger, Goldzycher, Planitzer, Kusa, Hanbury, and Neidhardt}]{pachinger-etal-2024-austrotox}
Pia Pachinger, Janis Goldzycher, Anna Planitzer, Wojciech Kusa, Allan Hanbury, and Julia Neidhardt. 2024.
\newblock \href {https://doi.org/10.18653/v1/2024.findings-acl.713} {{A}ustro{T}ox: A dataset for target-based {A}ustrian {G}erman offensive language detection}.
\newblock In \emph{Findings of the Association for Computational Linguistics: ACL 2024}, pages 11990--12001, Bangkok, Thailand. Association for Computational Linguistics.

\bibitem[{Pan et~al.(2024)Pan, Garc{\'\i}a-D{\'\i}az, and Valencia-Garc{\'\i}a}]{pan2024comparing}
Ronghao Pan, Jos{\'e}~Antonio Garc{\'\i}a-D{\'\i}az, and Rafael Valencia-Garc{\'\i}a. 2024.
\newblock Comparing fine-tuning, zero and few-shot strategies with large language models in hate speech detection in english.
\newblock \emph{CMES-Computer Modeling in Engineering \& Sciences}, 140(3).

\bibitem[{Pandey et~al.(2022)Pandey, Purohit, Castillo, and Shalin}]{pandey2022modeling}
Rahul Pandey, Hemant Purohit, Carlos Castillo, and Valerie~L Shalin. 2022.
\newblock Modeling and mitigating human annotation errors to design efficient stream processing systems with human-in-the-loop machine learning.
\newblock \emph{International Journal of Human-Computer Studies}, 160:102772.

\bibitem[{Park et~al.(2023{\natexlab{a}})Park, Kim, Park, and Park}]{park-etal-2023-k}
Chaewon Park, Soohwan Kim, Kyubyong Park, and Kunwoo Park. 2023{\natexlab{a}}.
\newblock \href {https://doi.org/10.18653/v1/2023.findings-emnlp.952} {K-{HATERS}: A hate speech detection corpus in {K}orean with target-specific ratings}.
\newblock In \emph{Findings of the Association for Computational Linguistics: EMNLP 2023}, pages 14264--14278, Singapore. Association for Computational Linguistics.

\bibitem[{Park et~al.(2023{\natexlab{b}})Park, Kim, Lee, Kang, Lee, Lee, and Lee}]{park-etal-2023-feel}
San-Hee Park, Kang-Min Kim, O-Joun Lee, Youjin Kang, Jaewon Lee, Su-Min Lee, and SangKeun Lee. 2023{\natexlab{b}}.
\newblock \href {https://doi.org/10.18653/v1/2023.findings-eacl.85} {{\textquotedblleft}why do {I} feel offended?{\textquotedblright} - {K}orean dataset for offensive language identification}.
\newblock In \emph{Findings of the Association for Computational Linguistics: EACL 2023}, pages 1142--1153, Dubrovnik, Croatia. Association for Computational Linguistics.

\bibitem[{Park et~al.(2024)Park, Kim, Jin, Park, and Han}]{park-etal-2024-predict}
Someen Park, Jaehoon Kim, Seungwan Jin, Sohyun Park, and Kyungsik Han. 2024.
\newblock \href {https://doi.org/10.18653/v1/2024.emnlp-main.1166} {{PREDICT}: Multi-agent-based debate simulation for generalized hate speech detection}.
\newblock In \emph{Proceedings of the 2024 Conference on Empirical Methods in Natural Language Processing}, pages 20963--20987, Miami, Florida, USA. Association for Computational Linguistics.

\bibitem[{Parvez(2025)}]{parvez-2025-chain}
Md~Rizwan Parvez. 2025.
\newblock \href {https://aclanthology.org/2025.knowledgenlp-1.21/} {Chain of evidences and evidence to generate: Prompting for context grounded and retrieval augmented reasoning}.
\newblock In \emph{Proceedings of the 4th International Workshop on Knowledge-Augmented Methods for Natural Language Processing}, pages 230--245, Albuquerque, New Mexic, USA. Association for Computational Linguistics.

\bibitem[{Potter et~al.(2024)Potter, Lai, Kim, Evans, and Song}]{potter-etal-2024-hidden}
Yujin Potter, Shiyang Lai, Junsol Kim, James Evans, and Dawn Song. 2024.
\newblock \href {https://doi.org/10.18653/v1/2024.emnlp-main.244} {Hidden persuaders: {LLM}s' political leaning and their influence on voters}.
\newblock In \emph{Proceedings of the 2024 Conference on Empirical Methods in Natural Language Processing}, pages 4244--4275, Miami, Florida, USA. Association for Computational Linguistics.

\bibitem[{Pradhan et~al.(2020)Pradhan, Chaturvedi, Tripathi, and Sharma}]{pradhan2020review}
Rahul Pradhan, Ankur Chaturvedi, Aprna Tripathi, and Dilip~Kumar Sharma. 2020.
\newblock A review on offensive language detection.
\newblock \emph{Advances in Data and Information Sciences: Proceedings of ICDIS 2019}, pages 433--439.

\bibitem[{Pujari et~al.(2024)Pujari, Wu, and Goldwasser}]{pujari-etal-2024-demand}
Rajkumar Pujari, Chengfei Wu, and Dan Goldwasser. 2024.
\newblock \href {https://doi.org/10.18653/v1/2024.emnlp-main.22} {{\textquotedblleft}we demand justice!{\textquotedblright}: Towards social context grounding of political texts}.
\newblock In \emph{Proceedings of the 2024 Conference on Empirical Methods in Natural Language Processing}, pages 362--372, Miami, Florida, USA. Association for Computational Linguistics.

\bibitem[{Ramos et~al.(2024)Ramos, Batista, Ribeiro, Fialho, Moro, Fonseca, Guerra, Carvalho, Marques, and Silva}]{ramos2024comprehensive}
Gil Ramos, Fernando Batista, Ricardo Ribeiro, Pedro Fialho, S{\'e}rgio Moro, Ant{\'o}nio Fonseca, Rita Guerra, Paula Carvalho, Catarina Marques, and Cl{\'a}udia Silva. 2024.
\newblock A comprehensive review on automatic hate speech detection in the age of the transformer.
\newblock \emph{Social Network Analysis and Mining}, 14(1):204.

\bibitem[{Rodr{\'\i}guez-Barroso et~al.(2024)Rodr{\'\i}guez-Barroso, C{\'a}mara, Collados, Luz{\'o}n, and Herrera}]{rodriguez2024federated}
Nuria Rodr{\'\i}guez-Barroso, Eugenio~Mart{\'\i}nez C{\'a}mara, Jose~Camacho Collados, M~Victoria Luz{\'o}n, and Francisco Herrera. 2024.
\newblock Federated learning for exploiting annotators’ disagreements in natural language processing.
\newblock \emph{Transactions of the Association for Computational Linguistics}, 12:630--648.

\bibitem[{Rosenthal et~al.(2021)Rosenthal, Atanasova, Karadzhov, Zampieri, and Nakov}]{rosenthal-etal-2021-solid}
Sara Rosenthal, Pepa Atanasova, Georgi Karadzhov, Marcos Zampieri, and Preslav Nakov. 2021.
\newblock \href {https://doi.org/10.18653/v1/2021.findings-acl.80} {{SOLID}: A large-scale semi-supervised dataset for offensive language identification}.
\newblock In \emph{Findings of the Association for Computational Linguistics: ACL-IJCNLP 2021}, pages 915--928, Online. Association for Computational Linguistics.

\bibitem[{Roy et~al.(2022)Roy, Bhawal, and Subalalitha}]{roy2022hate}
Pradeep~Kumar Roy, Snehaan Bhawal, and Chinnaudayar~Navaneethakrishnan Subalalitha. 2022.
\newblock Hate speech and offensive language detection in dravidian languages using deep ensemble framework.
\newblock \emph{Computer Speech \& Language}, 75:101386.

\bibitem[{Rozado(2024)}]{rozado2024political}
David Rozado. 2024.
\newblock The political preferences of llms.
\newblock \emph{PloS one}, 19(7):e0306621.

\bibitem[{Sainz et~al.(2023)Sainz, Campos, Garc{\'i}a-Ferrero, Etxaniz, de~Lacalle, and Agirre}]{sainz-etal-2023-nlp}
Oscar Sainz, Jon Campos, Iker Garc{\'i}a-Ferrero, Julen Etxaniz, Oier~Lopez de~Lacalle, and Eneko Agirre. 2023.
\newblock \href {https://doi.org/10.18653/v1/2023.findings-emnlp.722} {{NLP} evaluation in trouble: On the need to measure {LLM} data contamination for each benchmark}.
\newblock In \emph{Findings of the Association for Computational Linguistics: EMNLP 2023}, pages 10776--10787, Singapore. Association for Computational Linguistics.

\bibitem[{Schmidt and Wiegand(2017)}]{schmidt-wiegand-2017-survey}
Anna Schmidt and Michael Wiegand. 2017.
\newblock \href {https://doi.org/10.18653/v1/W17-1101} {A survey on hate speech detection using natural language processing}.
\newblock In \emph{Proceedings of the Fifth International Workshop on Natural Language Processing for Social Media}, pages 1--10, Valencia, Spain. Association for Computational Linguistics.

\bibitem[{Shi et~al.(2022)Shi, ValizadehAslani, Wang, Ren, Zhang, Hu, Zhao, and Liang}]{shi2022improving}
Yiwen Shi, Taha ValizadehAslani, Jing Wang, Ping Ren, Yi~Zhang, Meng Hu, Liang Zhao, and Hualou Liang. 2022.
\newblock Improving imbalanced learning by pre-finetuning with data augmentation.
\newblock In \emph{Fourth International Workshop on Learning with Imbalanced Domains: Theory and Applications}, pages 68--82. PMLR.

\bibitem[{Sun et~al.(2023)Sun, Li, Li, Wu, Guo, Zhang, and Wang}]{sun-etal-2023-text}
Xiaofei Sun, Xiaoya Li, Jiwei Li, Fei Wu, Shangwei Guo, Tianwei Zhang, and Guoyin Wang. 2023.
\newblock \href {https://doi.org/10.18653/v1/2023.findings-emnlp.603} {Text classification via large language models}.
\newblock In \emph{Findings of the Association for Computational Linguistics: EMNLP 2023}, pages 8990--9005, Singapore. Association for Computational Linguistics.

\bibitem[{Thapa et~al.(2024)Thapa, Rauniyar, Barkhordar, Veeramani, and Naseem}]{thapa-etal-2024-side}
Surendrabikram Thapa, Kritesh Rauniyar, Ehsan Barkhordar, Hariram Veeramani, and Usman Naseem. 2024.
\newblock \href {https://aclanthology.org/2024.alta-1.8/} {Which side are you on? investigating politico-economic bias in {N}epali language models}.
\newblock In \emph{Proceedings of the 22nd Annual Workshop of the Australasian Language Technology Association}, pages 104--117, Canberra, Australia. Association for Computational Linguistics.

\bibitem[{Tsoumou(2021)}]{tsoumou2021examination}
Jean~Mathieu Tsoumou. 2021.
\newblock An examination of verbal aggression in politically-motivated digital discourse.
\newblock \emph{International Journal of Social Media and Online Communities (IJSMOC)}, 13(2):22--43.

\bibitem[{Udawatta et~al.(2024)Udawatta, Udayangana, Gamage, Shekhar, and Ranathunga}]{udawatta2024use}
Pasindu Udawatta, Indunil Udayangana, Chathulanka Gamage, Ravi Shekhar, and Surangika Ranathunga. 2024.
\newblock Use of prompt-based learning for code-mixed and code-switched text classification.
\newblock \emph{World Wide Web}, 27(5):63.

\bibitem[{Vasist et~al.(2024)Vasist, Chatterjee, and Krishnan}]{vasist2024polarizing}
Pramukh~Nanjundaswamy Vasist, Debashis Chatterjee, and Satish Krishnan. 2024.
\newblock The polarizing impact of political disinformation and hate speech: a cross-country configural narrative.
\newblock \emph{Information Systems Frontiers}, 26(2):663--688.

\bibitem[{Voronov et~al.(2024)Voronov, Wolf, and Ryabinin}]{voronov-etal-2024-mind}
Anton Voronov, Lena Wolf, and Max Ryabinin. 2024.
\newblock \href {https://doi.org/10.18653/v1/2024.findings-acl.375} {Mind your format: Towards consistent evaluation of in-context learning improvements}.
\newblock In \emph{Findings of the Association for Computational Linguistics: ACL 2024}, pages 6287--6310, Bangkok, Thailand. Association for Computational Linguistics.

\bibitem[{Webb et~al.(2010)Webb, Sammut, Perlich, Horváth, Wrobel, Korb, Noble, Leslie, Lagoudakis, Quadrianto, Buntine, Getoor, Namata, Jin, Ting, Vijayakumar, Schaal, and De~Raedt}]{inbook}
Geoffrey Webb, Claude Sammut, Claudia Perlich, Tamás Horváth, Stefan Wrobel, Kevin Korb, William Noble, Christina Leslie, Michail Lagoudakis, Novi Quadrianto, Wray Buntine, Lise Getoor, Galileo Namata, Jiawei Jin, Jo-Anne Ting, Sethu Vijayakumar, Stefan Schaal, and Luc De~Raedt. 2010.
\newblock \href {https://doi.org/10.1007/978-0-387-30164-8_469} {\emph{Leave-One-Out Cross-Validation}}.

\bibitem[{Xiao et~al.(2024{\natexlab{a}})Xiao, Bouamor, and Zaghouani}]{xiao2024chinese}
Yunze Xiao, Houda Bouamor, and Wajdi Zaghouani. 2024{\natexlab{a}}.
\newblock Chinese offensive language detection: Current status and future directions.
\newblock \emph{arXiv preprint arXiv:2403.18314}.

\bibitem[{Xiao et~al.(2024{\natexlab{b}})Xiao, Hu, Choo, and Lee}]{xiao-etal-2024-toxicloakcn}
Yunze Xiao, Yujia Hu, Kenny Tsu~Wei Choo, and Roy Ka-Wei Lee. 2024{\natexlab{b}}.
\newblock \href {https://doi.org/10.18653/v1/2024.emnlp-main.345} {{T}oxi{C}loak{CN}: Evaluating robustness of offensive language detection in {C}hinese with cloaking perturbations}.
\newblock In \emph{Proceedings of the 2024 Conference on Empirical Methods in Natural Language Processing}, pages 6012--6025, Miami, Florida, USA. Association for Computational Linguistics.

\bibitem[{Yang et~al.(2023{\natexlab{a}})Yang, Zhang, Chen, Wang, and Kim}]{yang-etal-2023-prototype}
Weiyi Yang, Richong Zhang, Junfan Chen, Lihong Wang, and Jaein Kim. 2023{\natexlab{a}}.
\newblock \href {https://doi.org/10.18653/v1/2023.acl-long.904} {Prototype-guided pseudo labeling for semi-supervised text classification}.
\newblock In \emph{Proceedings of the 61st Annual Meeting of the Association for Computational Linguistics (Volume 1: Long Papers)}, pages 16369--16382, Toronto, Canada. Association for Computational Linguistics.

\bibitem[{Yang et~al.(2023{\natexlab{b}})Yang, Kim, Kim, Ho, Thorne, and Yun}]{yang-etal-2023-hare}
Yongjin Yang, Joonkee Kim, Yujin Kim, Namgyu Ho, James Thorne, and Se-Young Yun. 2023{\natexlab{b}}.
\newblock \href {https://doi.org/10.18653/v1/2023.findings-emnlp.365} {{HARE}: Explainable hate speech detection with step-by-step reasoning}.
\newblock In \emph{Findings of the Association for Computational Linguistics: EMNLP 2023}, pages 5490--5505, Singapore. Association for Computational Linguistics.

\bibitem[{Yi and Xia(2025)}]{yi2025irony}
Peiling Yi and Yuhan Xia. 2025.
\newblock Irony detection, reasoning and understanding in zero-shot learning.
\newblock \emph{arXiv preprint arXiv:2501.16884}.

\bibitem[{Yoo et~al.(2024)Yoo, Han, In, Jeon, Jeong, Kang, Kim, Kim, Kim, Kim et~al.}]{yoo2024hyperclova}
Kang~Min Yoo, Jaegeun Han, Sookyo In, Heewon Jeon, Jisu Jeong, Jaewook Kang, Hyunwook Kim, Kyung-Min Kim, Munhyong Kim, Sungju Kim, et~al. 2024.
\newblock Hyperclova x technical report.
\newblock \emph{arXiv preprint arXiv:2404.01954}.

\bibitem[{Yu et~al.(2024)Yu, Choi, and Kim}]{yu-etal-2024-dont}
Seunguk Yu, Juhwan Choi, and YoungBin Kim. 2024.
\newblock \href {https://doi.org/10.18653/v1/2024.findings-naacl.219} {Don`t be a fool: Pooling strategies in offensive language detection from user-intended adversarial attacks}.
\newblock In \emph{Findings of the Association for Computational Linguistics: NAACL 2024}, pages 3456--3467, Mexico City, Mexico. Association for Computational Linguistics.

\bibitem[{Zampieri et~al.(2019)Zampieri, Malmasi, Nakov, Rosenthal, Farra, and Kumar}]{zampieri-etal-2019-predicting}
Marcos Zampieri, Shervin Malmasi, Preslav Nakov, Sara Rosenthal, Noura Farra, and Ritesh Kumar. 2019.
\newblock \href {https://doi.org/10.18653/v1/N19-1144} {Predicting the type and target of offensive posts in social media}.
\newblock In \emph{Proceedings of the 2019 Conference of the North {A}merican Chapter of the Association for Computational Linguistics: Human Language Technologies, Volume 1 (Long and Short Papers)}, pages 1415--1420, Minneapolis, Minnesota. Association for Computational Linguistics.

\bibitem[{Zhang et~al.(2024)Zhang, Ladhak, Durmus, Liang, McKeown, and Hashimoto}]{zhang-etal-2024-benchmarking}
Tianyi Zhang, Faisal Ladhak, Esin Durmus, Percy Liang, Kathleen McKeown, and Tatsunori~B. Hashimoto. 2024.
\newblock \href {https://doi.org/10.1162/tacl_a_00632} {Benchmarking large language models for news summarization}.
\newblock \emph{Transactions of the Association for Computational Linguistics}, 12:39--57.

\bibitem[{Zhou et~al.(2023)Zhou, Zhu, Yerukola, Davidson, Hwang, Swayamdipta, and Sap}]{zhou-etal-2023-cobra}
Xuhui Zhou, Hao Zhu, Akhila Yerukola, Thomas Davidson, Jena~D. Hwang, Swabha Swayamdipta, and Maarten Sap. 2023.
\newblock \href {https://doi.org/10.18653/v1/2023.findings-acl.392} {{COBRA} frames: Contextual reasoning about effects and harms of offensive statements}.
\newblock In \emph{Findings of the Association for Computational Linguistics: ACL 2023}, pages 6294--6315, Toronto, Canada. Association for Computational Linguistics.

\bibitem[{Zou and Caragea(2023)}]{zou-caragea-2023-jointmatch}
Henry Zou and Cornelia Caragea. 2023.
\newblock \href {https://doi.org/10.18653/v1/2023.emnlp-main.451} {{J}oint{M}atch: A unified approach for diverse and collaborative pseudo-labeling to semi-supervised text classification}.
\newblock In \emph{Proceedings of the 2023 Conference on Empirical Methods in Natural Language Processing}, pages 7290--7301, Singapore. Association for Computational Linguistics.

\end{thebibliography}

\clearpage
\begin{table*}[t!]
\centering
\begin{adjustbox}{max width=\textwidth}
\begin{tabular}{l|rrrrrrrrrrrr|r}
\hline
                                                                                                 Topics      & \multicolumn{1}{c}{Jan}                                                            & \multicolumn{1}{c}{Feb}                                                            & \multicolumn{1}{c}{Mar}                                                            & \multicolumn{1}{c}{Apr}                                                            & \multicolumn{1}{c}{May}                                                            & \multicolumn{1}{c}{Jun}                                                            & \multicolumn{1}{c}{Jul}                                                            & \multicolumn{1}{c}{Aug}                                                            & \multicolumn{1}{c}{Sep}                                                            & \multicolumn{1}{c}{Oct}                                                            & \multicolumn{1}{c}{Nov}                                                            & \multicolumn{1}{c|}{Dec}                                                              & \multicolumn{1}{c}{Total}                                \\ \hline
\textit{Presidential Office}                                                                                    & \begin{tabular}[c]{@{}r@{}}(1,647 /\\ 118,499)\end{tabular}                         & \begin{tabular}[c]{@{}r@{}}(643 /\\ 40,031\end{tabular}                             & \begin{tabular}[c]{@{}r@{}}(544 /\\ 29,618)\end{tabular}                            & \begin{tabular}[c]{@{}r@{}}(911 /\\ 67,654)\end{tabular}                            & \begin{tabular}[c]{@{}r@{}}(949 /\\ 52,487)\end{tabular}                            & \begin{tabular}[c]{@{}r@{}}(388 /\\ 25,949)\end{tabular}                            & \begin{tabular}[c]{@{}r@{}}(797 /\\ 48,648)\end{tabular}                            & \begin{tabular}[c]{@{}r@{}}(1,072 /\\ 57,982)\end{tabular}                          & \begin{tabular}[c]{@{}r@{}}(1,040 /\\ 71,859)\end{tabular}                          & \begin{tabular}[c]{@{}r@{}}(738 /\\ 42,551)\end{tabular}                            & \begin{tabular}[c]{@{}r@{}}(1,119 /\\ 90,794)\end{tabular}                          & \begin{tabular}[c]{@{}r@{}}(3,425 /\\ 390,653)\end{tabular}                            & \begin{tabular}[c]{@{}r@{}}(13.2k /\\ 1.03m)\end{tabular} \\
\cellcolor[HTML]{EFEFEF}\begin{tabular}[c]{@{}l@{}}\textit{National Assembly}\\ \textit{/ Political Parties}\end{tabular} & \cellcolor[HTML]{EFEFEF}\begin{tabular}[c]{@{}r@{}}(2,679 /\\ 244,407)\end{tabular} & \cellcolor[HTML]{EFEFEF}\begin{tabular}[c]{@{}r@{}}(1,661 /\\ 130,922)\end{tabular} & \cellcolor[HTML]{EFEFEF}\begin{tabular}[c]{@{}r@{}}(2,158 /\\ 173,336)\end{tabular} & \cellcolor[HTML]{EFEFEF}\begin{tabular}[c]{@{}r@{}}(1,500 /\\ 142,668)\end{tabular} & \cellcolor[HTML]{EFEFEF}\begin{tabular}[c]{@{}r@{}}(2,437 /\\ 149,724)\end{tabular} & \cellcolor[HTML]{EFEFEF}\begin{tabular}[c]{@{}r@{}}(2,421 /\\ 167,195)\end{tabular} & \cellcolor[HTML]{EFEFEF}\begin{tabular}[c]{@{}r@{}}(2,875 /\\ 174,197)\end{tabular} & \cellcolor[HTML]{EFEFEF}\begin{tabular}[c]{@{}r@{}}(2,713 /\\ 146,288)\end{tabular} & \cellcolor[HTML]{EFEFEF}\begin{tabular}[c]{@{}r@{}}(2,148 /\\ 144,711)\end{tabular} & \cellcolor[HTML]{EFEFEF}\begin{tabular}[c]{@{}r@{}}(2,471 /\\ 138,529)\end{tabular} & \cellcolor[HTML]{EFEFEF}\begin{tabular}[c]{@{}r@{}}(2,622 /\\ 171,974)\end{tabular} & \cellcolor[HTML]{EFEFEF}\begin{tabular}[c]{@{}r@{}}(6,403 /\\ 653,247)\end{tabular}    & \begin{tabular}[c]{@{}r@{}}(32.0k /\\ 2.43m)\end{tabular} \\
\textit{North Korea}                                                                                            & \begin{tabular}[c]{@{}r@{}}(249 /\\ 15,340)\end{tabular}                            & \begin{tabular}[c]{@{}r@{}}(70 /\\ 1,977)\end{tabular}                              & \begin{tabular}[c]{@{}r@{}}(57 /\\ 2,333)\end{tabular}                              & \begin{tabular}[c]{@{}r@{}}(34 /\\ 1,243)\end{tabular}                              & \begin{tabular}[c]{@{}r@{}}(43 /\\ 1,738)\end{tabular}                              & \begin{tabular}[c]{@{}r@{}}(319 /\\ 15,744)\end{tabular}                            & \begin{tabular}[c]{@{}r@{}}(52 /\\ 1,023)\end{tabular}                              & \begin{tabular}[c]{@{}r@{}}(106 /\\ 4,340)\end{tabular}                             & \begin{tabular}[c]{@{}r@{}}(143 /\\ 4,566)\end{tabular}                             & \begin{tabular}[c]{@{}r@{}}(353 /\\ 16,341)\end{tabular}                            & \begin{tabular}[c]{@{}r@{}}(192 /\\ 3,943)\end{tabular}                             & \begin{tabular}[c]{@{}r@{}}(100 /\\ 7,907)\end{tabular}                                & \begin{tabular}[c]{@{}r@{}}(1.7k /\\ 76.4k)\end{tabular}  \\
\cellcolor[HTML]{EFEFEF}\textit{Administration}                                                                 & \cellcolor[HTML]{EFEFEF}\begin{tabular}[c]{@{}r@{}}(162 /\\ 4,708)\end{tabular}     & \cellcolor[HTML]{EFEFEF}\begin{tabular}[c]{@{}r@{}}(16 /\\ 359)\end{tabular}        & \cellcolor[HTML]{EFEFEF}\begin{tabular}[c]{@{}r@{}}(20 /\\ 1,465)\end{tabular}      & \cellcolor[HTML]{EFEFEF}\begin{tabular}[c]{@{}r@{}}(9 /\\ 146)\end{tabular}         & \cellcolor[HTML]{EFEFEF}\begin{tabular}[c]{@{}r@{}}(28 /\\ 1,235)\end{tabular}      & \cellcolor[HTML]{EFEFEF}\begin{tabular}[c]{@{}r@{}}(78 /\\ 3,219)\end{tabular}      & \cellcolor[HTML]{EFEFEF}\begin{tabular}[c]{@{}r@{}}(23 /\\ 356)\end{tabular}        & \cellcolor[HTML]{EFEFEF}\begin{tabular}[c]{@{}r@{}}(30 /\\ 1,922)\end{tabular}      & \cellcolor[HTML]{EFEFEF}\begin{tabular}[c]{@{}r@{}}(35 /\\ 989)\end{tabular}        & \cellcolor[HTML]{EFEFEF}\begin{tabular}[c]{@{}r@{}}(57 /\\ 2,189)\end{tabular}      & \cellcolor[HTML]{EFEFEF}\begin{tabular}[c]{@{}r@{}}(23 /\\ 882)\end{tabular}        & \cellcolor[HTML]{EFEFEF}\begin{tabular}[c]{@{}r@{}}(112 /\\ 7,667)\end{tabular}        & \begin{tabular}[c]{@{}r@{}}(0.5k /\\ 25.1k)\end{tabular}  \\
\begin{tabular}[c]{@{}l@{}}\textit{National Defense}\\ \textit{/ Foreign Affairs}\end{tabular}                            & \begin{tabular}[c]{@{}r@{}}(130 /\\ 5,126)\end{tabular}                             & \begin{tabular}[c]{@{}r@{}}(53 /\\ 1,775)\end{tabular}                              & \begin{tabular}[c]{@{}r@{}}(82 /\\ 2,829)\end{tabular}                              & \begin{tabular}[c]{@{}r@{}}(64 /\\ 1,495)\end{tabular}                              & \begin{tabular}[c]{@{}r@{}}(221 /\\ 9,347)\end{tabular}                             & \begin{tabular}[c]{@{}r@{}}(208 /\\ 8,656)\end{tabular}                             & \begin{tabular}[c]{@{}r@{}}(96 /\\ 2,773)\end{tabular}                              & \begin{tabular}[c]{@{}r@{}}(173 /\\ 5,613)\end{tabular}                             & \begin{tabular}[c]{@{}r@{}}(155 /\\ 7,325)\end{tabular}                             & \begin{tabular}[c]{@{}r@{}}(110 /\\ 3,917)\end{tabular}                             & \begin{tabular}[c]{@{}r@{}}(178 /\\ 8,483)\end{tabular}                             & \begin{tabular}[c]{@{}r@{}}(607 /\\ 35,451)\end{tabular}                               & \begin{tabular}[c]{@{}r@{}}(2.0k /\\ 92.7k)\end{tabular}  \\
\cellcolor[HTML]{EFEFEF}\textit{General Politics}                                                               & \cellcolor[HTML]{EFEFEF}\begin{tabular}[c]{@{}r@{}}(7,054 /\\ 672,500)\end{tabular} & \cellcolor[HTML]{EFEFEF}\begin{tabular}[c]{@{}r@{}}(5,734 /\\ 489,416)\end{tabular} & \cellcolor[HTML]{EFEFEF}\begin{tabular}[c]{@{}r@{}}(7,086 /\\ 572,271)\end{tabular} & \cellcolor[HTML]{EFEFEF}\begin{tabular}[c]{@{}r@{}}(4,849 /\\ 495,238)\end{tabular} & \cellcolor[HTML]{EFEFEF}\begin{tabular}[c]{@{}r@{}}(961 /\\ 72,292)\end{tabular}    & \cellcolor[HTML]{EFEFEF}\begin{tabular}[c]{@{}r@{}}(3,272 /\\ 253,428)\end{tabular} & \cellcolor[HTML]{EFEFEF}\begin{tabular}[c]{@{}r@{}}(4,427 /\\ 311,983)\end{tabular} & \cellcolor[HTML]{EFEFEF}\begin{tabular}[c]{@{}r@{}}(5,001 /\\ 329,824)\end{tabular} & \cellcolor[HTML]{EFEFEF}\begin{tabular}[c]{@{}r@{}}(4,362 /\\ 336,104)\end{tabular} & \cellcolor[HTML]{EFEFEF}\begin{tabular}[c]{@{}r@{}}(4,398 /\\ 273,979)\end{tabular} & \cellcolor[HTML]{EFEFEF}\begin{tabular}[c]{@{}r@{}}(5,203 /\\ 389,716)\end{tabular} & \cellcolor[HTML]{EFEFEF}\begin{tabular}[c]{@{}r@{}}(12,655 /\\ 1,422,883)\end{tabular} & \begin{tabular}[c]{@{}r@{}}(65.0k /\\ 5.61m)\end{tabular} \\ \hline
Total                                                                                                  & \begin{tabular}[c]{@{}r@{}}(11.9k /\\ 1.06m)\end{tabular}                           & \begin{tabular}[c]{@{}r@{}}(8.1k /\\ 0.66m)\end{tabular}                            & \begin{tabular}[c]{@{}r@{}}(9.9k /\\ 0.78m)\end{tabular}                            & \begin{tabular}[c]{@{}r@{}}(7.3k /\\ 0.70m)\end{tabular}                            & \begin{tabular}[c]{@{}r@{}}(4.6k /\\ 0.28m)\end{tabular}                            & \begin{tabular}[c]{@{}r@{}}(6.6k /\\ 0.47m)\end{tabular}                            & \begin{tabular}[c]{@{}r@{}}(8.2k /\\ 0.53m)\end{tabular}                            & \begin{tabular}[c]{@{}r@{}}(9.0k /\\ 0.54m)\end{tabular}                            & \begin{tabular}[c]{@{}r@{}}(7.8k /\\ 0.56m)\end{tabular}                            & \begin{tabular}[c]{@{}r@{}}(8.1k /\\ 0.47m)\end{tabular}                            & \begin{tabular}[c]{@{}r@{}}(9.3k /\\ 0.66m)\end{tabular}                            & \begin{tabular}[c]{@{}r@{}}(23.3k /\\ 2.51m)\end{tabular}                              & \begin{tabular}[c]{@{}r@{}}(\textbf{114k} /\\ \textbf{9.28m})\end{tabular}  \\ \hline
\end{tabular}
\end{adjustbox}
\caption{Overview of the total number of political news articles and corresponding comments collected from the Naver in 2024. The values on the left and right represent the number of articles and comments, respectively.}
\label{table_politicalko_statistics}
\end{table*}

\appendix

\section{Further Details in\protect\linebreak\hspace*{\parindent}Constructing \textbf{\textit{PoliticalK.O}}{\Large\texttwemoji{boxing glove}}}
\label{appendix_a}

Since the target news platform has recorded the highest user traffic in South Korea, the initial volume of collected comments was exceptionally large. To include only comments of appropriate length, we examined the length distribution across the five Korean offensive language datasets used in this study. The analysis showed that the shortest and longest 10\% of comments averaged 11.8 and 85.2 characters. Accordingly, we filtered the comments ranging from 12 to 85 characters in length. The resulting distributions of news articles and comments are provided in Table~\ref{table_politicalko_statistics}, revealing clear variations in volume by topic and publication date\footnote{Notably, December 2024 saw a sharp increase in political news coverage and comment activity, largely due to public debate over the presidential declaration of martial law.}.

%\section{Further Details in\\\;\;\;\;\;\;Offensive Language Judgments}
\section{Further Details in\protect\linebreak\hspace*{\parindent}Offensive Language Judgments}
\label{appendix_b}

\subsection{Supervised Ensemble Judgment}
\label{appendix_b1}

\textbf{Selection of PLMs} We selected RoBERTa\footnote{\url{https://huggingface.co/FacebookAI/xlm-roberta-base}} for its multilingual configuration and consistent performance on Korean tasks. We also referred to previous study indicating that KcBERT\footnote{\url{https://huggingface.co/beomi/kcbert-base}}, pre-trained on noisy text such as online comments, is effective in classifying similarly noisy inputs~\cite{yu-etal-2024-dont}.

\textbf{Statistics from prior datasets} All offensive language datasets used in this study showed label imbalance, with detailed statistics provided in Table~\ref{table_prior_datasets_statistics}. We observed that nearly all datasets contained a significantly higher proportion of non-offensive labels. To mitigate the potential impact of issue on unseen comments, we re-split the train, valid, and test sets of each dataset into an 8:1:1 ratio, ensuring an equal distribution of offensive and non-offensive labels across all subsets.

\textbf{Fine-tuning setup \& results} We trained 12 layers of each model with a dropout rate of 0.1, optimizing with AdamW at learning rates of \{1e-5, 2e-5, 3e-5\} for 5 epochs\footnote{When we extended to 10 epochs in an effort to improve performance, this resulted in overfitting to the training data.}. To select the best model, we evaluated both the checkpoint with the lowest validation loss and the final epoch, and chose the one that yielded the highest F1 score.

The best-performing results for each offensive language dataset are presented in Table~\ref{table_pej_plm_eval}. We observed that the optimal conditions for achieving the highest score varied across the datasets. To ensure robust prediction on unseen comments in \textbf{\textit{PoliticalK.O}}{\Large\texttwemoji{boxing glove}}, we selected the best-trained model for each dataset as best\_model$_i$ in Algorithm~\ref{algorithm_sej}.

\begin{table}[t!]
\centering
\small
\begin{adjustbox}{max width=\columnwidth}
\begin{tabular}{l|c|c|c}
\hline
Prior Datasets         & Collection Period         & \textit{Label Distribution} & Volume \\ \hline
K-Haters & July -- August 2021       & 18.06 : \textbf{81.94}      & 192.1k \\
KODOLI   & October -- December 2020  & 34.62 : \textbf{65.38}      & 38.5k  \\
KoLD     & March 2020 -- March 2022  & 4.95 : \textbf{95.05}       & 40.4k  \\
K-MHaS   & January 2018 -- June 2020 & 45.65 : \textbf{54.35}      & 109.6k \\
UnSmile  & January 2019 -- June 2020 & \textbf{65.06} : 24.94      & 18.7k  \\ \hline
\textbf{\textit{PoliticalK.O}}{\Large\texttwemoji{boxing glove}} & \textbf{January -- December 2024} & in Table~\ref{table_ld_and_lo}                  & \textbf{9.28m}  \\ \hline
\end{tabular}
\end{adjustbox}
\caption{Comparison of existing and proposed offensive language datasets by key characteristics The label distribution values (left : right) represent the proportion of (offensive : non-offensive) labels.}
\label{table_prior_datasets_statistics}
\end{table}

\begin{table}[t!]
\centering
\small
\begin{adjustbox}{max width=\columnwidth}
\begin{tabular}{l|ccc|ccc}
\hline
\multirow{3}{*}{\begin{tabular}[c]{@{}l@{}}Prior\\ Datasets\end{tabular}}         & \multicolumn{3}{c|}{RoBERTa}                                                                    & \multicolumn{3}{c}{KcBERT}                                                                               \\ \cline{2-7} 
         & \multicolumn{1}{c|}{\begin{tabular}[c]{@{}c@{}}Best\\ (epoch, lr)\end{tabular}} & \textit{Acc}   & \textit{F1}    & \multicolumn{1}{c|}{\begin{tabular}[c]{@{}c@{}}Best\\ (epoch, lr)\end{tabular}} & \textit{Acc}   & \textit{F1}             \\ \hline
K-Haters & \multicolumn{1}{c|}{5, 1e-5}                                                    & 81.68 & 75.72 & \multicolumn{1}{c|}{5, 1e-5}                                                    & 83.49 & \textbf{77.76} \\
KODOLI   & \multicolumn{1}{c|}{5, 1e-5}                                                    & 89.38 & 88.07 & \multicolumn{1}{c|}{5, 2e-5}                                                    & 91.22 & \textbf{90.24} \\
KoLD     & \multicolumn{1}{c|}{5, 2e-5}                                                    & 80.21 & 80.20 & \multicolumn{1}{c|}{4, 1e-5}                                                    & 83.84 & \textbf{83.84} \\
K-MHaS   & \multicolumn{1}{c|}{4, 1e-5}                                                    & 87.51 & 87.35 & \multicolumn{1}{c|}{5, 1e-5}                                                    & 89.90 & \textbf{89.82} \\
UnSmile  & \multicolumn{1}{c|}{4, 2e-5}                                                    & 84.16 & 77.86 & \multicolumn{1}{c|}{4, 2e-5}                                                    & 87.57 & \textbf{83.30} \\ \hline
\end{tabular}
\end{adjustbox}
\caption{Performance on the re-split test sets of five prior offensive language datasets under optimal conditions.}
\label{table_pej_plm_eval}
\end{table}

\subsection{Prompt-variants Ensemble Judgment}
\label{appendix_b2}

\textbf{Selection of LLMs} The LLMs used in this study were pre-trained from scratch on large-scale Korean datasets, without leveraging weights from existing models. Exaone\footnote{\url{https://huggingface.co/LGAI-EXAONE/EXAONE-3.5-7.8B-Instruct}} was released in December 2024, followed by Trillion\footnote{\url{https://huggingface.co/trillionlabs/Trillion-7B-preview}} and HyperclovaX\footnote{\url{https://huggingface.co/naver-hyperclovax/HyperCLOVAX-SEED-Text-Instruct-1.5B}} in March and April 2025, respectively. We set the temperature to 0 and utilized the vLLM library to ensure efficient inference~\cite{kwon2023efficient}.

\textbf{Details in \textit{Summ} prompt} We provided a three-sentence summary of each corresponding article, including its title and content. The summarization followed a zero-shot setting informed by prior studies~\cite{chhabra-etal-2024-revisiting, zhang-etal-2024-benchmarking}, with temperature 0.3 and top\_p 0.5 identified as optimal parameters for generating informative outputs~\cite{houamegni2025evaluating}.

We conducted an evaluation to assess the quality of the summaries. Following the prior studies~\cite{jia-etal-2023-zero, gao2023human}, each summary was rated on a 1-5 Likert scale based on its factual alignment with the source article. We implemented a self-assessment framework where the model evaluated the consistency of its own outputs.

The evaluation results of article summaries are reported in Table~\ref{table_summaries_eval}. Both Trillion and HyperclovaX scored in the upper 4-point range, while Exaone scored slightly lower but still close to 4. Although HyperclovaX exhibited a higher standard deviation, this is reasonable given its 1.5B parameters compared to the 7B of the other models.

\textbf{Details in \textit{FewShots} prompt} Given the potential impact of labeled samples on the few-shot learning~\cite{bragg2021flex}, we define the pseudo-label annotation criteria used in this study. Comments containing critical language without explicit profanity were not labeled as offensive if they reflected personal opinions rather than targeting specific individual or groups\footnote{While sensitive language can affect labeling decisions from a broader perspective, we focused on the factual context of political discourse, where certain expressions of personal stance are considered acceptable.}. We sampled user comments from news articles published in January and February 2025, ensuring no overlap with the dataset collection period. To mitigate the impact of noise on label predictions~\cite{chen2023automatic}, we corrected the grammar of the sample comments, allowing for a more focused assessment of their offensiveness.

The samples for each topic are provided in Tables~\ref{table_samples_president}-\ref{table_samples_politics}. Expressions like \textit{brainwashed sheeple} and \textit{blind and deaf} in Table~\ref{table_samples_president} reflect extreme language. Similarly, phrases such as \textit{They're so `brilliant' it's terrifying for our future.} in Table~\ref{table_samples_politics} may seem innocuous but convey strong sarcasm targeting a specific group. In contrast, comments that expressed criticism without targeting someone were labeled as non-offensive.

\subsection{Multi-debate Reasoning Judgment}
\label{appendix_b3}

\textbf{Model selection} This judgment requires five interconnected inferences per comment, and due to the scale of our dataset, applying it across all models would be prohibitively time-consuming. To identify the most suitable model, we evaluated the detection performance of three LLMs using a \textit{Vanilla} prompt on existing offensive language datasets. The results are provided in Table~\ref{table_mrj_llm_eval}.

Trillion achieved an average F1 score of 78.06 and was selected as the primary model for our main analysis. However, its performance remained below that of the fine-tuned PLM under optimal conditions (Table~\ref{table_pej_plm_eval}). This underscores the limitations of even recent LLMs in zero-shot settings and the need for tailored detection methods.

\begin{table}[t!]
\centering
\small
\begin{adjustbox}{max width=0.95\columnwidth}
\begin{tabular}{l|c|c|c}
\hline
Topics                                                                               & Exaone       & Trillion     & HyperclovaX \\ \hline
\textit{Presidential Office}                                                            & (3.96, 0.41) & (4.75, 0.48) & (4.86, 0.64)           \\ 
\rowcolor[HTML]{EFEFEF} \begin{tabular}[c]{@{}l@{}}\textit{National Assembly}\\ \textit{/ Political Parties}\end{tabular} & (3.98, 0.31) & (4.73, 0.48) & (4.84, 0.72)           \\
\textit{North Korea}                                                                    & (4.02, 0.29) & (4.76, 0.45) & (4.85, 0.68)           \\
\rowcolor[HTML]{EFEFEF} \textit{Administration}                                                                 & (3.98, 0.39) & (4.73, 0.53) & (4.82, 0.78)           \\
\begin{tabular}[c]{@{}l@{}}\textit{National Defense}\\ \textit{/ Foreign Affairs}\end{tabular}    & (3.97, 0.37) & (4.69, 0.55) & (4.86, 0.67)           \\
\rowcolor[HTML]{EFEFEF} \textit{General Politics}                                                               & (4.00, 0.31) & (4.75, 0.45) & (4.86, 0.66)           \\ \hline
\end{tabular}
\end{adjustbox}
\caption{Summary consistency scores (mean, standard deviation) based on self-assessments by each model.}
\label{table_summaries_eval}
\end{table}

\begin{table}[t!]
\centering
\small
\begin{adjustbox}{max width=0.95\columnwidth}
\begin{tabular}{l|cc|cc|cc}
\hline
\multirow{2}{*}{\begin{tabular}[c]{@{}l@{}}Prior\\ Datasets\end{tabular}}         & \multicolumn{2}{c|}{Exaone} & \multicolumn{2}{c|}{Trillion} & \multicolumn{2}{c}{HyperclovaX} \\ \cline{2-7} 
         & \textit{Acc}           & \textit{F1}          & \textit{Acc}            & \textit{F1}           & \textit{Acc}             & \textit{F1}            \\ \hline
K-Haters & 82.05            & 73.66          & 78.04             & 73.96           & 77.59              & 69.17            \\
KODOLI   & 76.09            & 75.54          & 81.62             & 80.08           & 76.09              & 74.89            \\
KoLD     & 80.95            & 80.81          & 77.96             & 77.72           & 74.79              & 74.75            \\
K-MHaS   & 66.32            & 64.32          & 77.75             & 77.64           & 70.32              & 69.92            \\
UnSmile  & 85.92            & 78.98          & 84.90             & 80.90           & 79.46              & 73.49            \\ \hline
Avg      & 78.26            & 74.66          & 80.05             & \textbf{78.06}           & 75.65              & 72.44            \\ \hline
\end{tabular}
\end{adjustbox}
\caption{Performance on the re-split test sets of five prior offensive language datasets under different LLMs.}
\label{table_mrj_llm_eval}
\end{table}

\clearpage
\begin{table*}[t!]
\centering
\small
\begin{adjustbox}{max width=\textwidth}
\begin{tabular}{l|c}
\hline
\multicolumn{1}{c|}{Sample Comments} &
  Label \\ \hline
\begin{tabular}[c]{@{}l@{}}계엄은 대통령의 권한이다. 계엄을 내란이란 떠들고 범죄라고 떠드는 놈들은 공산주의 좌파놈들과 공산좌파언론에 속은 개돼지들 틀림없다.\\ \;\;\;\;(\textit{Martial law is the President's prerogative.}\\ \;\;\;\;\;\textit{\textbf{Anyone screaming that it's rebellion or calling it a crime is nothing but a brainwashed sheeple, duped by the commie-left and their media puppets.}})\end{tabular} &
  Offensive \\ \hline
\begin{tabular}[c]{@{}l@{}}윤석열이 1년 넘게 계엄을 입에 담았다는데 용산 종자들이 시각장애우나 청각장애우도 아니고 몰랐을 리가 있겠느냐\\ 전원 출국금지 시키고 내란 공범 여부를 철저하게 조사해야 한다.\\ 저런 것들은 없어도 나라 돌아가는데 아무 지장없다. 시사평론가 장모씨 말로는 요즘은 출근해서 유튜브나 본다고 하더만.\\ \;\;\;\;(\textit{They say Yoon's been talking about martial law for over a year — \textbf{do you really think the cronies in Yongsan were blind and deaf the whole time?}}\\ \;\;\;\;\;\textit{They all need travel bans and a full investigation into whether they were co-conspirators in treason. \textbf{Honestly, we’d be better off without these people.}}\\ \;\;\;\;\;\textit{According to political commentator Mr. Jang, \textbf{they don’t even work anymore — just sit around watching YouTube all day.}})\end{tabular} &
  Offensive \\ \hline
\rowcolor[HTML]{EFEFEF} \begin{tabular}[c]{@{}l@{}}윤석열 계엄이나 전부 사직으로 막았어야지, 헌법재판관 임명 반대가 더 중요해?\\ \;\;\;\;(\textit{Yoon Suk-yeol should’ve stopped this with martial law or by forcing everyone to resign.}\\ \;\;\;\;\;\textit{You're telling me opposing a Constitutional Court appointment was more important?})\end{tabular} &
  Non-offensive \\ \hline
\rowcolor[HTML]{EFEFEF} \begin{tabular}[c]{@{}l@{}}세금이 아깝다. 사표를 수리하고, 내란 동조 관련 혐의는 철저히 수사해야 한다.\\ 비서실장이 계엄을 몰랐을 리 없다. 정진석도 한덕수처럼 윤석열에게 코 꿰어 어쩔 수 없이 지지하는 것처럼 보인다.\\ \;\;\;\;(\textit{What a waste of taxpayers' money. Their resignations should be accepted, and there must be a thorough investigation into charges of aiding an insurrection.}\\ \;\;\;\;\;\textit{There's no way the Chief of Staff didn' t know about the martial law plans.}\\ \;\;\;\;\;\textit{Jeong Jin-seok is starting to look just like Han Duck-soo — hooked by Yoon and left with no choice but to support him.})\end{tabular} &
  Non-offensive \\ \hline
\end{tabular}
\end{adjustbox}
\caption{Few-shot samples used for the comments under the topic \textit{Presidential Office}.}
\label{table_samples_president}
\end{table*}

\begin{table*}[t!]
\centering
\small
\begin{adjustbox}{max width=\textwidth}
\begin{tabular}{l|c}
\hline
\multicolumn{1}{c|}{Sample Comments} &
  Label \\ \hline
\begin{tabular}[c]{@{}l@{}}그러니까 니가 공화당 따위 대표나 하고 앉아 있는 거야.\\ \;\;\;\;(\textit{\textbf{That’s exactly why you’re stuck as the so-called leader of some wannabe Republican knockoff.}})\end{tabular} &
  Offensive \\ \hline
\begin{tabular}[c]{@{}l@{}}국힘이 잘해서 지지하는게 아니고 리재명이 싫어서 할 수 없이 견제하는거다. 정치인 다 정신 차려야 된다.\\ \;\;\;\;(\textit{People aren’t supporting the PPP because they’re doing a great job}\\ \;\;\;\;\;\textit{— \textbf{they just can’t stand Lee Jae-myung, so they’re backing them as a necessary evil.} Every politician needs to snap out of it.})\end{tabular} &
  Offensive \\ \hline
\rowcolor[HTML]{EFEFEF} \begin{tabular}[c]{@{}l@{}}저 인간, 전에 국회에서 난리치려다가 끌려나간 인간이었지 아마?\\ \;\;\;\;(\textit{Wasn't that guy the one who tried to cause a scene in the National Assembly and ended up getting dragged out?})\end{tabular} &
  Non-offensive \\ \hline
\rowcolor[HTML]{EFEFEF} \begin{tabular}[c]{@{}l@{}}맞음 국민의 힘 지지율은 국민의힘 지지율이라기 보다 윤석열 대통령 복귀를 위한 지지율임 고로 배신때린 자들은 대권 생각마셈.\\ \;\;\;\;(\textit{Exactly. The support ratings for the People Power Party aren’t really about the party — they’re basically a proxy for backing President Yoon’s return.}\\ \;\;\;\;\;\textit{So anyone who stabbed him in the back shouldn’t even dream of running for president.})\end{tabular} &
  Non-offensive \\ \hline
\end{tabular}
\end{adjustbox}
\caption{Few-shot samples used for the comments under the topic \textit{National Assembly / Political Parties}.}
\label{table_samples_national_assembly}
\end{table*}

\begin{table*}[t!]
\centering
\small
\begin{adjustbox}{max width=\textwidth}
\begin{tabular}{l|c}
\hline
\multicolumn{1}{c|}{Sample Comments} &
  Label \\ \hline
\begin{tabular}[c]{@{}l@{}}공산주의자와는 대화가 안된다. 우리가 핵무기 10개만 있으면 대화하자고 나올 거다. 트럼프는 순진한가나? 바보인가?\\ \;\;\;\;(\textit{\textbf{There’s no talking with communists.} But if we had just ten nukes, they'd be the ones begging to negotiate.}\\ \;\;\;\;\;\textit{\textbf{What, was Trump just being naive? Or straight-up stupid?}})\end{tabular} &
  Offensive \\ \hline
\begin{tabular}[c]{@{}l@{}}공산주의 놈들 신났네. 역시 사상과 이념이 공산주의라 트럼프가 만나준다니까? 아주 좋아 죽네. 공산주의 놈들아 니들은 사상이 문제야 사상이.\\ 김정은 한테 욕한번 해봐라. 못하지? 김정은이 그리 좋으냐? 우리 동포들이 지금도 굶어서 죽는다. 이 공산주의 놈들아.\\ \;\;\;\;(\textit{\textbf{The commie bastards must be thrilled.} Of course they’re loving it — Trump’s actually agreeing to meet with them. \textbf{Typical of their twisted ideology.}}\\ \;\;\;\;\;\textit{\textbf{Hey, why don’t you try cursing out Kim Jong-un once? Can’t do it, can you? You love that dictator so much?}}\\ \;\;\;\;\;\textit{Our fellow Koreans are still starving to death, and \textbf{you commie scum sit there grinning. It’s your rotten ideology — that’s the real problem.}})\end{tabular} &
  Offensive \\ \hline
\rowcolor[HTML]{EFEFEF} \begin{tabular}[c]{@{}l@{}}이제 정은이가 콧대 높아져서 남한은 쳐다도 안보겠네 잘됐네. 우리 우파는 이럴수록 더 뭉칩시다.\\ \;\;\;\;(\textit{Now that little Jung-eun’s gotten all high and mighty, he probably won’t even bother looking South anymore.}\\ \;\;\;\;\;\textit{Good riddance. This is exactly when we conservatives need to stand even stronger together.})\end{tabular} &
  Non-offensive \\ \hline 
\rowcolor[HTML]{EFEFEF} \begin{tabular}[c]{@{}l@{}}헛물켜지마라. 자고로 미국은 결정적일때 우리를 버렸다. 비단 미국만이 아니다. 지금 우크라이나를 봐라. 힘없는 나라는 평화도 없다.\\ 물론 미국에게 고마운 것도 많지만 우리가 이뻐서 도와주는 것만은 아님을 누구나 다 아는 사실이다.\\ 진영 상대를 적으로만 대하지 말고 인정하고 빨리 국력을 키워야한다.\\ \;\;\;\;(\textit{Don’t get your hopes up. History shows the U.S. always bails when it really matters. And it’s not just the U.S.}\\ \;\;\;\;\;\textit{— look at Ukraine. Weak countries don’t get peace. Sure, we owe the U.S. a lot, but let’s not kid ourselves — they’re not helping us out of love.}\\ \;\;\;\;\;\textit{Everyone knows that. Instead of treating political opponents like enemies, we should acknowledge reality and focus on building national strength.})\end{tabular} &
  Non-offensive \\ \hline
\end{tabular}
\end{adjustbox}
\caption{Few-shot samples used for the comments under the topic \textit{North Korea}.}
\label{table_samples_north_korea}
\end{table*}

\clearpage
\begin{table*}[t!]
\centering
\small
\begin{adjustbox}{max width=\textwidth}
\begin{tabular}{l|c}
\hline
\multicolumn{1}{c|}{Sample Comments} &
  Label \\ \hline
\begin{tabular}[c]{@{}l@{}}생각이라는게 있긴 한거임? 국무위원들을 그냥 지졸개 정도로 생각했겠지 모지란 것이.\\ \;\;\;\;(\textit{\textbf{Do you even have the ability to think? You probably saw the cabinet ministers as nothing more than clueless flunkies — typical of someone that dim.}})\end{tabular} &
  Offensive \\ \hline
\begin{tabular}[c]{@{}l@{}}계엄자체가 정상이 아닌데 무슨 윤수괴한테 정상적인 것을 기대하냐. 친일도 계몽이라는 놈한테 무슨 정상적인 절차가 있었겠냐.\\ \;\;\;\;(\textit{Martial law itself is completely abnormal — \textbf{why would anyone expect anything normal from Yoon the Tyrant?}}\\ \;\;\;\;\;\textit{\textbf{The guy who called Japanese colonialism 'enlightenment'} — you think he cares about following due process?})\end{tabular} &
  Offensive \\ \hline
\rowcolor[HTML]{EFEFEF} \begin{tabular}[c]{@{}l@{}}어찌되었든 계엄은 잘못된 거라고 생각한다. 윤석열씨는 내려오는게 맞다고 생각한다.\\ \;\;\;\;(\textit{Regardless of the details, I believe declaring martial law was wrong. Yoon Suk-yeol should step down.})\end{tabular} &
  Non-offensive \\ \hline
\rowcolor[HTML]{EFEFEF} \begin{tabular}[c]{@{}l@{}}계엄은 대통령과 국방장관이 계획했고, 결심권자는 대통령이고, 국무회의는 형식상의 절차였고, 국무위원은 의결권이 없고, 의견발언권만 있다.\\ 계엄은 합법이고, 설령 위법성이 있다하여도, 대통령은 사소한 법위반으로 기소되지 않는 특권이 있다. 나는 대통령의 비상계엄을 지지한다.\\ \;\;\;\;(\textit{Martial law was planned by the President and the Defense Minister. The final authority rests solely with the President.}\\ \;\;\;\;\;\textit{The cabinet meeting was just a formality — ministers don’t have voting power, only the right to express opinions.}\\ \;\;\;\;\;\textit{Martial law is legal. And even if some legal issues arise, the President is constitutionally immune from prosecution over minor violations.}\\ \;\;\;\;\;\textit{I fully support the President’s emergency martial law declaration.})\end{tabular} &
  Non-offensive \\ \hline
\end{tabular}
\end{adjustbox}
\caption{Few-shot samples used for the comments under the topic \textit{Administration}.}
\label{table_samples_administration}
\end{table*}

\begin{table*}[t!]
\centering
\small
\begin{adjustbox}{max width=\textwidth}
\begin{tabular}{l|c}
\hline
\multicolumn{1}{c|}{Sample Comments} &
  Label \\ \hline
\begin{tabular}[c]{@{}l@{}}간첩이 득실대고 있는데 민주건달들은 무슨 짓을 하고 있는 거냐?\\ \;\;\;\;(\textit{While spies are crawling all over the place, \textbf{what the hell are these Demo-thugs even doing?}})\end{tabular} &
  Offensive \\ \hline
\begin{tabular}[c]{@{}l@{}}1억 받고 中에 블랙요원 신상 넘긴 군무원? 혹시 더듬이당 당원인지 한 번 까봐라.\\ \;\;\;\;(\textit{A military official who sold out a black agent's identity to China for a hundred million won? \textbf{Someone check if he’s a member of the Stammering Party.}})\end{tabular} &
  Offensive \\ \hline
\rowcolor[HTML]{EFEFEF} \begin{tabular}[c]{@{}l@{}}국가분열행위, 방산비리인데 재산 몰수 후, 사형이 맞지않나?\\ \;\;\;\;(\textit{This is an act of treason and a massive defense corruption scandal — shouldn't the punishment be asset seizure and the death penalty?})\end{tabular} &
  Non-offensive \\ \hline
\rowcolor[HTML]{EFEFEF} \begin{tabular}[c]{@{}l@{}}정말 나라 기강이 다 무너지고 있다. 사태가 이 지경인데 민주당은 왜 간첩법 거부하는가? 정말 이해 할 수 없다.\\ \;\;\;\;(\textit{The entire moral fabric of the nation is collapsing.}\\ \;\;\;\;\;\textit{Things are in total chaos, and yet the Democratic Party is rejecting the Anti-Spy Law? I just can’t make sense of it.})\end{tabular} &
  Non-offensive \\ \hline
\end{tabular}
\end{adjustbox}
\caption{Few-shot samples used for the comments under the topic \textit{National Defense / Foreign Affairs}.}
\label{table_samples_national_defense}
\end{table*}

\begin{table*}[t!]
\centering
\small
\begin{adjustbox}{max width=\textwidth}
\begin{tabular}{l|c}
\hline
\multicolumn{1}{c|}{Sample Comments} &
  Label \\ \hline
\begin{tabular}[c]{@{}l@{}}경기지사면 지사답게 도정에만 좀 제발 좀 신경써라. 뭐 임방한다고 나와서 떠드냐.\\ \;\;\;\;(\textit{If you're the governor of Gyeonggi Province, then act like one and focus on running the province.}\\ \;\;\;\;\;\textit{\textbf{Why the hell are you out here livestreaming and ranting like an influencer?}})\end{tabular} &
  Offensive \\ \hline
\begin{tabular}[c]{@{}l@{}}돈만 찍어내서 뿌려대고 서민 경제 살렸다며,\\ 자신들이 지운 빚은 지들 죽은 뒤의 미래 세대에 물려주는 존나 유능한 진보가 민주당. 너무 유능해서 미래가 걱정된다.\\ \;\;\;\;(\textit{\textbf{The oh-so-`competent progressives' of the Democratic Party keep printing money},}\\ \;\;\;\;\;\textit{\textbf{throwing it around, and patting themselves on the back for `saving' the economy}}\\ \;\;\;\;\;\textit{— \textbf{all while dumping the debt they created on future generations after they’re long gone. They're so `brilliant' it’s terrifying for our future.}})\end{tabular} &
  Offensive \\ \hline
\rowcolor[HTML]{EFEFEF} \begin{tabular}[c]{@{}l@{}}보수냐 진보냐 그게 중요한 것이 아니야. 이성적으로 판단하고 세상을 사는 것이 답이야. 이념 탓만 하니 나라 꼴이 이러지.\\ \;\;\;\;(\textit{It’s not about being conservative or progressive — what really matters is thinking rationally and living with common sense.}\\ \;\;\;\;\;\textit{This obsession with ideology is exactly why the country’s such a mess.})\end{tabular} &
  Non-offensive \\ \hline 
\rowcolor[HTML]{EFEFEF} \begin{tabular}[c]{@{}l@{}}유능한 진보건, 중도 보수건 일 잘하고 국민 삶 윤택해지면 최고다! 청년일자리, 미중 패권전쟁에서 국익 지키는 것,\\ 출산율 저하로 미래 대한민국 존립성 위태로움 극복도 큰과제. 또한 대북 리스크 줄이는 것도 코리아 디스카운트 줄이는 큰 과제일것이다!\\ \;\;\;\;(\textit{Whether it's competent progressives or moderate conservatives, what matters is results — improving people’s lives.}\\ \;\;\;\;\;\textit{We need to create jobs for young people, protect our national interests in the U.S.-China power struggle,}\\ \;\;\;\;\;\textit{and tackle the existential threat posed by declining birthrates. Reducing the North Korea risk is also key to ending the `Korea Discount' once and for all.})\end{tabular} &
  Non-offensive \\ \hline
\end{tabular}
\end{adjustbox}
\caption{Few-shot samples used for the comments under the topic \textit{General Politics}.}
\label{table_samples_politics}
\end{table*}

\clearpage

\section{Further Experimental Results in\protect\linebreak\hspace*{\parindent}Three Main Judgments}
\label{appendix_c}

\subsection{Evaluating Judgments on Ground Truth}
\label{appendix_c1}

We conducted human annotation on a subset to evaluate the performance of each judgment against actual ground truth, involving five university graduates who labeled 20 comments across six topics, totaling 120 comments each. Annotators were provided with labeling criteria detailed in Appendix~\ref{appendix_b2} and example annotations from Tables~\ref{table_samples_president}-\ref{table_samples_politics}. We then aggregated their results via majority voting to derive final labels used as ground truth. The Krippendorff's $\alpha$ was 0.62, indicating moderate agreement among human raters.

The evaluation results of each judgment on a human-annotated subset are presented in Table~\ref{table_ground_truth_eval_3j}. While \texttt{PEJ} achieved the highest performance in our main analysis, \texttt{SEJ} slightly outperformed it when evaluated against ground truth. This suggests that the offensive language datasets used in \texttt{SEJ} may align closely with the labeling criteria adopted in our study. Although these results are based on only 0.001\% of the dataset due to the limited scale of human annotation, we observed a consistent trend of higher scores across all judgments when evaluated using ground trust\textsuperscript{\dag}. This implies that follow-up analyses should consider the potential overestimation introduced by judgment-based automated evaluations. 

\begin{table}[h!]
\centering
\small
\begin{adjustbox}{max width=0.9\columnwidth}
\begin{tabular}{ll|c|ccc}
\hline
\multicolumn{2}{l|}{Each Judgment} &
  \textit{Acc} &
  \textit{P} &
  \textit{R} &
  \textit{F1} \\ \hline
\multicolumn{1}{l|}{} &
  ground truth &
  65.83 &
  74.02 &
  73.16 &
  65.81 \\
\multicolumn{1}{l|}{\multirow{-2}{*}{\texttt{SEJ}}} &
  \cellcolor[HTML]{EFEFEF}ground trust\textsuperscript{\dag} &
  \cellcolor[HTML]{EFEFEF}86.66 &
  \cellcolor[HTML]{EFEFEF}80.15 &
  \cellcolor[HTML]{EFEFEF}90.10 &
  \cellcolor[HTML]{EFEFEF}82.95 \\ \hline
\multicolumn{1}{l|}{} &
  ground truth &
  55.00 &
  71.87 &
  65.38 &
  53.96 \\
\multicolumn{1}{l|}{\multirow{-2}{*}{\texttt{PEJ}}} &
  \cellcolor[HTML]{EFEFEF}ground trust\textsuperscript{\dag} &
  \cellcolor[HTML]{EFEFEF}97.50 &
  \cellcolor[HTML]{EFEFEF}96.87 &
  \cellcolor[HTML]{EFEFEF}95.47 &
  \cellcolor[HTML]{EFEFEF}96.15 \\ \hline
\multicolumn{1}{l|}{} &
  ground truth &
  51.66 &
  71.00 &
  62.82 &
  49.98 \\
\multicolumn{1}{l|}{\multirow{-2}{*}{\texttt{MRJ}}} &
  \cellcolor[HTML]{EFEFEF}ground trust\textsuperscript{\dag} &
  \cellcolor[HTML]{EFEFEF}90.83 &
  \cellcolor[HTML]{EFEFEF}88.50 &
  \cellcolor[HTML]{EFEFEF}82.42 &
  \cellcolor[HTML]{EFEFEF}84.95 \\ \hline
\end{tabular}
\end{adjustbox}
\caption{Evaluation of each judgment on a human-annotated subset using both ground truth and ground trust\textsuperscript{\dag} for all topics.}
\label{table_ground_truth_eval_3j}
\end{table}

\subsection{MRJ on Other Models}
\label{appendix_c2}

We applied \texttt{MRJ} to the subset corresponding to May 2024 that contains \textit{0.28 million} comments, using Exaone and HyperclovaX with the analysis presented in the Table~\ref{table_mrj_ld_and_lo}. Both Exaone and Trillion predicted a higher proportion of offensive labels, whereas HyperclovaX showed a lower rate, around 50\%. While HyperclovaX exhibited a comparable overlap ratio to other models with \texttt{SEJ}, its ratio with \texttt{PEJ} was notably lower, suggesting that prompt- and debate-level consistency may be more limited in smaller models.

\begin{table}[h!]
\centering
\small
\begin{adjustbox}{max width=0.95\columnwidth}
\begin{tabular}{l|c|cc}
\hline
\multirow{2}{*}{Models} &
  \multirow{2}{*}{\begin{tabular}[c]{@{}c@{}}\textit{Label}\\ \textit{Distribution}\end{tabular}} &
  \begin{tabular}[c]{@{}c@{}}\texttt{PEJ}\\ $\rightleftharpoons$ \texttt{MRJ}\end{tabular} &
  \begin{tabular}[c]{@{}c@{}}\texttt{MRJ}\\ $\rightleftharpoons$ \texttt{SEJ}\end{tabular} \\ \cline{3-4} 
            &               & \multicolumn{2}{c}{\textit{Pairwise Label Overlap Ratio}} \\ \hline
\rowcolor[HTML]{EFEFEF} Exaone      & 85.39 : 14.60 & 87.00                   & 69.97                  \\ \hline
Trillion    & 81.70 : 18.29 & 83.91                   & 69.85                  \\ \hline
\rowcolor[HTML]{EFEFEF} HyperclovaX & 50.39 : 49.60 & 64.56                   & 66.73                  \\ \hline
\end{tabular}
\end{adjustbox}
\caption{Label distribution and pairwise label overlap ratio using \texttt{MRJ} on a subset across all models and topics.}
\label{table_mrj_ld_and_lo}
\end{table}

\subsection{Case Study}
\label{appendix_c3}

The agreement examples for each judgment are provided in Table~\ref{table_case_study}. Cases uniquely labeled as non-offensive by \texttt{SEJ} often lacked explicit profanity, but the LLM-based approach appears to have classified them as offensive depending on context. For \texttt{PEJ}, comments conveying implicit hostility toward \textit{hardcore supporters} were still labeled non-offensive. Lastly, non-offensive judgment by \texttt{MRJ} typically involved personal political opinions that, upon an objective perspective, exhibited no clear signs of offensiveness. These cases illustrate the varying interpretations that can arise from different judgment criteria, even for the same comment.

\begin{table}[h!]
\centering
\small
\begin{adjustbox}{max width=\columnwidth}
\begin{tabular}{ccc}
\hline
\multicolumn{1}{c|}{\texttt{SEJ}} &
  \multicolumn{1}{c|}{\texttt{PEJ}} &
  \texttt{MRJ} \\ \hline
\multicolumn{1}{c|}{\textbf{\textit{Non-off}}} &
  \multicolumn{1}{c|}{\textit{Off}} &
  \textit{Off} \\ \hline
\multicolumn{3}{c}{\cellcolor[HTML]{EFEFEF}\begin{tabular}[c]{@{}c@{}}도대체 국민 70\%가 찬성하는 김건희 특검은\\ 거부하면서 공정과 상식은 어디에 있는지?\\ (\textit{So they reject a special investigation into Kim Keon-hee}\\ \textit{that 70\% of the public wants, do they still}\\ \textit{have the nerve to talk about fairness and common sense?})\end{tabular}} \\ \hline
\multicolumn{1}{c|}{\texttt{SEJ}} &
  \multicolumn{1}{c|}{\texttt{PEJ}} &
  \texttt{MRJ} \\ \hline
\multicolumn{1}{c|}{\textit{Off}} &
  \multicolumn{1}{c|}{\textbf{\textit{Non-off}}} &
  \textit{Off} \\ \hline
\multicolumn{3}{c}{\cellcolor[HTML]{EFEFEF}\begin{tabular}[c]{@{}c@{}}여야 강성들만 보고 정치하면 망하는 지름길이다.\\ (\textit{When you let only the hardcore supporters}\\ \textit{from both sides shape your politics,}\\ \textit{you're setting yourself up for failure.})\end{tabular}} \\ \hline
\multicolumn{1}{c|}{\texttt{SEJ}} &
  \multicolumn{1}{c|}{\texttt{PEJ}} &
  \texttt{MRJ} \\ \hline
\multicolumn{1}{c|}{\textit{Off}} &
  \multicolumn{1}{c|}{\textit{Off}} &
  \textbf{\textit{Non-off}} \\ \hline
\multicolumn{3}{c}{\cellcolor[HTML]{EFEFEF}\begin{tabular}[c]{@{}c@{}}자기 당과 대통령 망하라고 여론전 하는 인간이 여기 있네.\\ (\textit{Funny how he's out there pushing narrative}\\ \textit{that make it look like}\\ \textit{he wants his own party and president to fail.})\end{tabular}} \\ \hline
\end{tabular}
\end{adjustbox}
\caption{A case study uniquely identified as non-offensive by each judgment. \textit{Off} and \textit{Non-off} denote offensive and non-offensive, respectively.}
\label{table_case_study}
\end{table}

%\onecolumn
\clearpage

\section{Prompt Constructions}
\label{appendix_d}

\vspace{1cm}
\begin{itemize}
    \item The following prompts are used in our \textbf{prompt-variants ensemble judgment} (\texttt{PEJ}).\vspace*{0.5em}
    \fbox{
      \begin{minipage}{0.85\columnwidth}
      \small
        \#\# \textit{system}\\
        Your task is to classify whether the given comment on Korean political news articles is offensive.\\
        \\
        \#\# \textit{user}\\
        \textit{If}\;\;\(\textcolor{blue}{Defn}\)\{\\
        \hspace*{0.5em}Offensive language refers to words or expressions intended to insult, harm, or belittle individuals or groups. Especially in the context of Korean political discussions, this includes:\\
        \hspace*{0.5em}- Insulting, derogatory, or combative language that may be directed at political figures, parties, or ideologies.\\
        \hspace*{0.5em}- Discriminatory language based on political affiliation, race, religion, gender, or other personal attributes.\\
        \hspace*{0.5em}- Sarcastic or harmful humor that may target political beliefs, parties, or individuals in a demeaning way.\\
        \hspace*{0.5em}- Incitement of violence or hatred against political opponents, minority groups, or societal institutions.\\
        \hspace*{0.5em}- Misinformation or harmful narratives that spread unverified or damaging political views, especially in a divisive or inflammatory manner.\\
        \}\\
        \textit{If}\;\;\(\textcolor{blue}{Summ}\)\{\\
        \hspace*{0.5em}The following sentences are a summary of the original article on which the comment was posted.\\
        \\
        \hspace*{0.5em}Summary: \{\textit{summary of the source article}\}\\
        \}\\
        \textit{If}\;\;\(\textcolor{blue}{FewShots}\)\{\\
        \hspace*{0.5em}You may refer to the following examples.\\
        \\
        \hspace*{0.5em}Comment: \{\textit{sample comment under the same topic with target comment}\}\\
        \\
        \hspace*{0.5em}Label: \{\textit{sample pseudo-label under the same topic with target comment}\}\\
        \hspace*{0.5em}...\\
        \hspace*{0.5em}Comment: \{\textit{sample comment under the same topic with target comment}\}\\
        \\
        \hspace*{0.5em}Label: \{\textit{sample pseudo-label under the same topic with target comment}\}\\
        \}\\
        \textit{If}\;\;\(\textcolor{blue}{Vanilla}\)\{\\
        \hspace*{0.5em}Please classify whether the comment is offensive or not. Respond with only "yes" or "no", without any explanations.\\
        \\
        \hspace*{0.5em}Comment: \{\textit{target comment}\}\\
        \\
        \hspace*{0.5em}Label:\\
        \}
      \end{minipage}
    }
    \newpage\vspace*{3em}
    \item\justifying The following prompts are used in the \textit{Summ} prompt for summary generation and evaluation to each model.\par\vspace*{0.5em}
    \noindent\fbox{
        \begin{minipage}{0.85\columnwidth}  
        \small  
        \textcolor{blue}{\#\# Summary generation}\\
        \hspace*{0.5em}\#\# \textit{system}\\
        \hspace*{0.5em}Your task is to provide consistent summaries of Korean political news articles.\\
        \\
        \hspace*{0.5em}\#\# \textit{user}\\
        \hspace*{0.5em}Based on the following title and content, please summarize the news article in three sentences.\\
        \\
        \hspace*{0.5em}Title: \{\textit{title of the source article}\}\\
        \\
        \hspace*{0.5em}Content: \{\textit{content of the source article}\}\\
        \\
        \hspace*{0.5em}Summary:\\
        \\
        \textcolor{blue}{\#\# Summary evaluation}\\
        \hspace*{0.5em}\#\# \textit{system}\\
        \hspace*{0.5em}Your task is to evaluate how well the given summary reflects the content of Korean political news articles.\\
        \\
        \hspace*{0.5em}\#\# \textit{user}\\
        \hspace*{0.5em}Please score the following summary for its consistency with the corresponding article on a scale from 1 to 5, without any explanations. Consistency refers to how much of the information in the summary is actually present in the source article. A score of 5 means that all statements in the summary are fully supported by the article, while a score of 1 means the summary does not reflect the article at all.\\
        \\
        \hspace*{0.5em}Summary: \{\textit{summarized article}\}\\
        \\
        \hspace*{0.5em}Source Title: \{\textit{title of the source article}\}\\
        \\
        \hspace*{0.5em}Source Content: \{\textit{content of the source article}\}\\
        \\
        \hspace*{0.5em}Score:
        \end{minipage}
    }
    \newpage\vspace*{2em}
    \item The following prompts are used in our \textbf{multi-debate reasoning judgment} (\texttt{MRJ}).\vspace*{0.5em}
    \fbox{
        \begin{minipage}{0.85\columnwidth}  
        \small
        \textcolor{blue}{\#\# Persona alignment for LLM$_{\textit{O}}$ (LLM$_{\textit{N}}$})\\
        \hspace*{0.5em}\#\# \textit{system}\\
        \hspace*{0.5em}You argue that the given comment from Korean political news is offensive (non-offensive), and provide a clear justification based on linguistic and contextual cues.\\
        \\
        \hspace*{0.5em}\#\# \textit{user}\\
        \hspace*{0.5em}Please classify whether the comment is offensive or not. Respond with only "yes" or "no", without any explanations.\\
        \\
        \hspace*{0.5em}Comment: \{\textit{target comment}\}\\
        \\
        \hspace*{0.5em}Label:\\
        \\
        \hspace*{0.5em}\#\# \textit{assistant}\\
        \hspace*{0.5em}\{\textit{yes} (\textit{no})\}\\
        \\
        \textcolor{blue}{\#\# Rationale generation}\\
        \hspace*{0.5em}\#\# \textit{user}\\
        \hspace*{0.5em}You think the comment is offensive (non-offensive).\\
        \\
        \hspace*{0.5em}The following sentences are a summary of the original article on which the comment was posted.\\
        \\
        \hspace*{0.5em}Summary: \{\textit{summary of the source article}\}\\
        \\
        Based on the context, given a reason why you think the comment is offensive (non-offensive). Please do so in one sentence.\\
        \\
        \hspace*{0.5em}Reason:
        \end{minipage}
    }\vspace*{0.5em}
    \fbox{
        \begin{minipage}{0.85\columnwidth}  
        \small
        \textcolor{blue}{\#\# Discuss on each side}\\
        \hspace*{0.5em}\#\# \textit{system}\\
        \hspace*{0.5em}You argue that the given comment from Korean political news is offensive (non-offensive), and provide a clear justification based on linguistic and contextual cues.\\
        \\
        \hspace*{0.5em}\#\# \textit{user}\\
        \hspace*{0.5em}You have to consider the given comment.\\
        \\
        \hspace*{0.5em}Comment: \{\textit{target comment}\}\\
        \\
        \hspace*{0.5em}Your argument about this comment:\\
        \hspace*{0.5em}\{\textit{rationale$_\textit{O}$} (\textit{rationale$_\textit{N}$})\}\\
        \\
        \hspace*{0.5em}The non-offensive (offensive) side's argument about this comment:\\
        \hspace*{0.5em}\{\textit{rationale$_\textit{N}$} (\textit{rationale$_\textit{O}$})\}\\
        \\
        \hspace*{0.5em}Based on your argument, agree or rebut the non-offensive (offensive) sides' argument and explain your reason. Please do so in one sentence.\\
        \\
        \hspace*{0.5em}Stance:
        \end{minipage}
    }
    \newpage\vspace*{5.19em}
    \noindent\fbox{
        \begin{minipage}{0.85\columnwidth}  
        \small
        \textcolor{blue}{\#\# Final judgment}\\
        \hspace*{0.5em}\#\# \textit{system}\\
        \hspace*{0.5em}Your task is classify whether the given comment on Korean political news articles is offensive. You should consider both sides, offensive and non-offensive, fairly to maintain a balanced perspective.\\
        \\
        \hspace*{0.5em}\#\# \textit{user}\\
        \hspace*{0.5em}As a judge, assess the debaters' arguments and stances based on the following criteria, "How well they capture the non-offensiveness or offensiveness of the comment". Consider both sides fairly to maintain a balanced perspective and make a broad judgment.\\
        \\
        \hspace*{0.5em}The offensive side's argument: \{\textit{rationale$_\textit{O}$}\}\\
        \hspace*{0.5em}The non-offensive side's argument: \{\textit{rationale$_\textit{N}$}\}\\
        \\
        \hspace*{0.5em}The offensive side's stance: \{\textit{stance$_\textit{O}$}\}\\
        \hspace*{0.5em}The non-offensive side's stance: \{\textit{stance$_\textit{N}$}\}\\
        \\
        \hspace*{0.5em}Please classify whether the comment is offensive or not. Respond with only "yes" or "no", without any explanations.\\
        \\
        \hspace*{0.5em}Comment: \{\textit{target comment}\}\\
        \\
        \hspace*{0.5em}Label:
        \end{minipage}
    }
\end{itemize}

\end{document}